%% file: emnlp2023.tex
\pdfoutput=1

\documentclass[11pt]{article}

\usepackage[]{EMNLP2023}

\usepackage{times}
\usepackage{latexsym}

\usepackage[T1]{fontenc}

\usepackage[utf8]{inputenc}

\usepackage{microtype}

\usepackage{inconsolata}

\usepackage{booktabs}
\usepackage{threeparttable}
\usepackage{multirow}
\usepackage{amsfonts}
\usepackage{amsmath}
\usepackage{graphicx}
\usepackage[linesnumbered,ruled,vlined]{algorithm2e}
\usepackage{soul}  
\usepackage{enumitem} 
\usepackage{colortbl} 
\usepackage{pifont}

\newcommand \footnoteONLYtext[1]
{
	\let \mybackup \thefootnote
	\let \thefootnote \relax
	\footnotetext{#1}
	\let \thefootnote \mybackup
	\let \mybackup \imareallyundefinedcommand
}

\usepackage{longtable} 
\usepackage{CJKutf8}
\usepackage{float}

\title{\fontsize{14}{2}\selectfont Uncovering Limitations of Large Language Models \\[1.0ex] in Information Seeking from Tables}

\author{\fontsize{11}{2}\selectfont Chaoxu Pang$^{1,2}$,\,Yixuan Cao$^{1,2*}$,\,Chunhao Yang$^{4}$,\,Ping Luo$^{1,2,3*}$ \\
  \fontsize{11}{2}\selectfont  $^1$Key Lab of Intelligent Information Processing of Chinese Academy of Sciences (CAS) \\
  \fontsize{11}{2}\selectfont Institute of Computing Technology, CAS, Beijing 100190, China \\
  \fontsize{11}{2}\selectfont $^2$University of Chinese Academy of Sciences, Beijing 100049, China \\
  \fontsize{11}{2}\selectfont $^3$Peng Cheng Laboratory, Shenzhen 518066, China \\
  \fontsize{11}{2}\selectfont $^4$Harbin Engineering University, Harbin 150001, China \\
  \fontsize{9}{2}\selectfont \texttt{\{pangchaoxu21b,caoyixuan,luop\}@ict.ac.cn, doublehappy@hrbeu.edu.cn} \\}

\begin{document}
\begin{CJK*}{UTF8}{gbsn} 

\maketitle
\footnoteONLYtext{\textsuperscript{*}Corresponding author: Yixuan Cao and Ping Luo.}
\vspace{2cm}

\input{abstract}

\input{introduction2}
\input{tabis_bench}
\input{experiment2}
\input{related_work}

\input{conclusion}
\input{limitation}
\bibliography{anthology,custom}
\bibliographystyle{acl_natbib}

\input{appendix}
\end{CJK*}

\end{document}

%% file: abstract.tex
\begin{abstract}
Tables are recognized for their high information density and widespread usage, serving as essential sources of information. 
Seeking information from tables (TIS) is a crucial capability for Large Language Models (LLMs), serving as the foundation of knowledge-based Q\&A systems.
However, this field presently suffers from an absence of thorough and reliable evaluation. This paper introduces a more reliable benchmark for \textbf{Ta}ble \textbf{I}nformation \textbf{S}eeking (TabIS). To avoid the unreliable evaluation caused by text similarity-based metrics, TabIS adopts a single-choice question format (with two options per question) instead of a text generation format. 
We establish an effective pipeline for generating options, ensuring their  difficulty and quality. 
Experiments conducted on 12 LLMs reveal that while the performance of GPT-4-turbo is marginally satisfactory, both other proprietary and open-source models perform inadequately. Further analysis shows that LLMs exhibit a poor understanding of table structures, and struggle to balance between TIS performance and robustness against pseudo-relevant tables (common in retrieval-augmented systems). These findings uncover the limitations and potential challenges of LLMs in seeking information from tables. We release our data and code to facilitate further research in this field.\footnote{\url{https://github.com/coszero/TabIS}}

\end{abstract}

%% file: introduction2.tex
\section{Introduction}

\begin{figure}[t]
\centering
\includegraphics[width=0.48\textwidth]{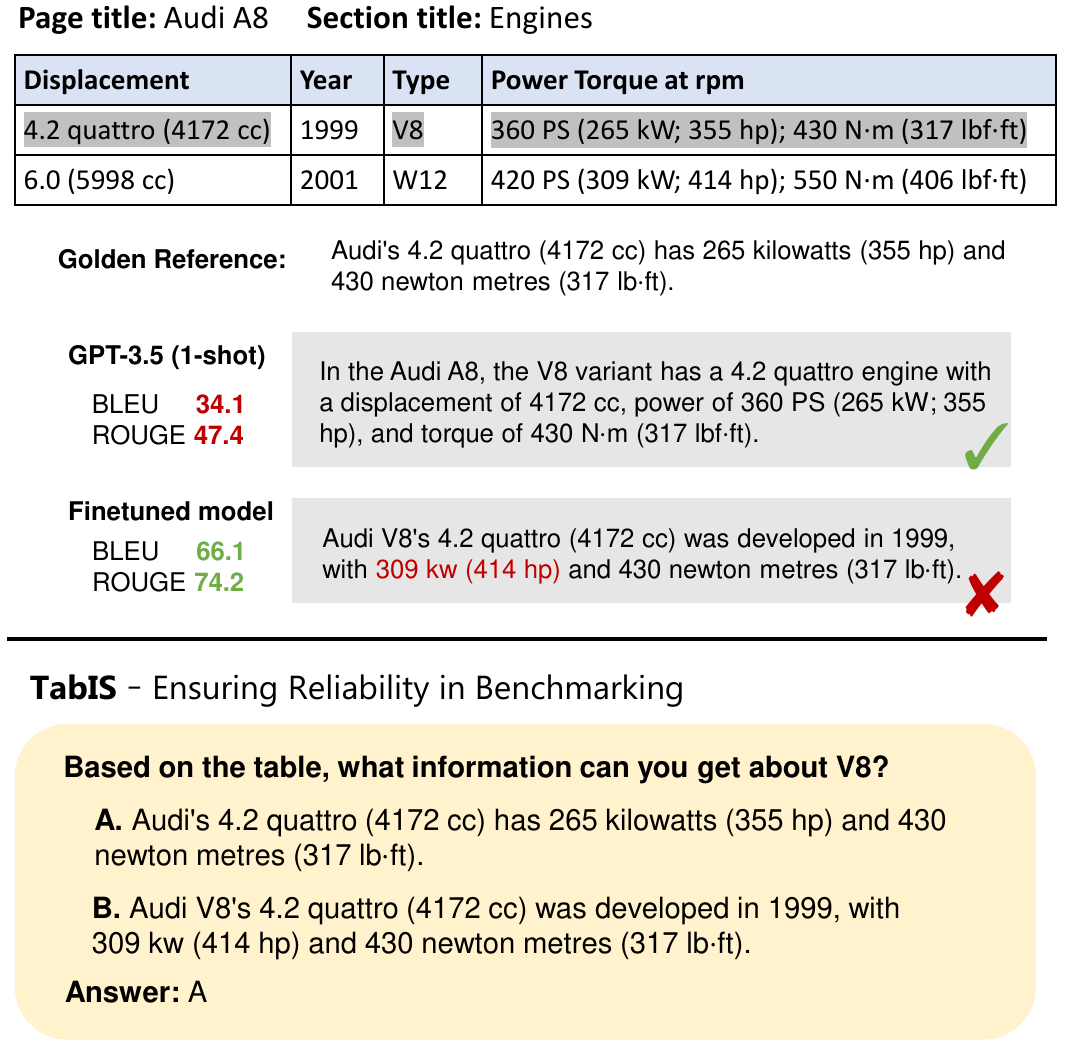}
\caption{Above: A simplified table-to-text generation example illustrating the unreliable evaluation issue. Higher values on surface-level metrics like BLEU and ROUGE do not guarantee better results. Target cells are highlighted. Below: Our benchmark presented in a single-choice format.}
\label{fig:eval_example}
\end{figure}

Tables are widespread and rich sources of information across the web and in various documents. Statistics show that the number of tables on internet web pages has reached several hundred million~\cite{lehmberg2016large}; in the corporate environment, the number of tables in Excel-like spreadsheet files has exceeded 115 million~\cite{wang2020structure}. Precisely seeking relevant information from tables is crucial for a wide array of real-world applications, including financial analysis, scientific research, etc. Recently, the remarkable advancements of Large Language Models (LLMs)~\citep{gpt3, palm, gpt4, llama2, gemini-pro} have transformed the approach of information retrieval, moving from fetching specific passages to directly providing answers. However, the effectiveness of LLMs in seeking information from tables remains underexplored.

Some efforts have been made to evaluate the capabilities of LLMs in table information seeking (TIS), but there are unreliable evaluation issues with the used evaluation metrics. 
Previous studies~\citep{invest_ttg} mainly use table-to-text generation (TTG) as a test bench to assess the TIS abilities of LLMs. TTG aims at transforming complex tabular data into comprehensible descriptions tailored to users’ information seeking needs.
The evaluation relies heavily on surface-level metrics such as BLEU~\citep{bleu} and ROUGE~\citep{rouge}, or on metrics based on model predictions such as NLI-Acc~\citep{nli_acc}.
Given that LLM responses can greatly differ in style from reference answers, using these metrics can lead to inconsistent and unreliable evaluations. An example of this issue is illustrated in Figure~\ref{fig:eval_example} where a fine-tuned model's incorrect description receives higher BLEU/ROUGE scores than the correct output from GPT-3.5. This discrepancy may occur because GPT-3.5, without being fine-tuned on this specific dataset, might not mimic the style of the reference response. 

To provide a more reliable evaluation, this paper introduces a new benchmark for \textbf{Tab}le \textbf{I}nformation \textbf{S}eeking (TabIS). 
We design our benchmark using a single-choice question format, motivated by popular benchmarks like MMLU~\citep{mmlu} and BBH~\citep{bbh}, which utilize this format to offer a reliable and widely accepted evaluation of LLMs.
We convert TTG datasets like ToTTo~\citep{totto} and Hitab~\citep{hitab} into this format so that the results can be simply and reliably evaluated.
A challenge during curating this benchmark is to generate high-quality options for single-choice questions. Initially, the original data's answer could serve as the correct option. So we need to generate a \emph{deceptive} wrong option. If the generated option is too simple, e.g. with obvious logical errors or unrelated to the table content, the benchmark will be too easy and fail to test LLMs' capabilities. To address this, we devised three prompting-based methods: Modify-Input, Modify-Output, and Exam-Judge (detailed in Section \ref{sec:sis_option_generation}) for generating wrong options.
These methods together produced a variety of deceptive options. The manually verified accuracy rate of our generated data exceeds 92\%. We also noted that the Exam-Judge method we proposed generated more challenging questions, which may be used for future dataset construction.

Leveraging the high-quality options, TabIS encompasses three practical scenarios with increasing difficulty for table information seeking: (1) basic TIS derived from TTG (B-TIS), (2) TIS that emphasizes structural understanding (SU-TIS), i.e. when directed to a specific table area with position information (row and column), and (3) TIS from multiple tables (M-TIS), i.e. when confronted additional pseudo-relevant tables. These scenarios reflect common challenges in real-world applications, such as chatbots and retrieval-augmented systems. 

While previous studies ~\citep{invest_ttg} that test on the basic TIS setting with unreliable metrics demonstrate the superiority of LLMs, TabIS reveals the limitations and potential challenges of LLMs in table information seeking as follows.

\begin{itemize}[leftmargin=0.5cm]
    \item \textbf{Most LLMs show suboptimal TIS performance, especially in complex TIS scenarios and when handling tables with rich hierarchies.} Experiments on 12 representative LLMs show that only GPT-4-turbo attained an 85.7\% accuracy on average (random guess would be 50\% accuracy). The top-performing 70B open-source model achieved 74.4\%, with the rest falling in the 50-60\% range. 
    \item \textbf{LLMs exhibit a poor understanding of table structures, with accuracy fluctuating across different cell positions.} Surprisingly, we find that LLMs perform almost at random levels in basic lookup tasks, such as repeating content in a specific row. This highlights the substantial challenges in real-world SU-TIS scenarios, where models struggle to pinpoint the target table area using only positional cues.
    \item \textbf{LLMs struggle to balance between TIS performance and robustness against pseudo-relevant tables, especially for open-source models.} This indicates a great challenge for LLMs in retrieval-augmented generation scenarios.
\end{itemize}


Finally, we fine-tune \emph{Llama2-13b-chat} on our weakly-supervised training dataset and find that while fine-tuning can significantly improve TIS performance, boosting from 55.5 to 73.2, it still lags behind GPT-4-turbo, which has not been specifically fine-tuned. This indicates that the proposed benchmark is non-trivial, calling for further investigations and improvement in this field.




%% file: tabis_bench.tex
\section{TabIS Benchmark}\label{tabis_bench}

We curated a benchmark \emph{TabIS} to investigate the table information seeking capabilities of LLMs. 


We use table-to-text generation (TTG) datasets as the original data source in our benchmark. The task of TTG is that, given a table and a set of selected cells $(T, C)$, produce a one-sentence description of the cells, and the annotated description is called ``reference'' $R$. We transform TTG into a single-choice question with two options for objective and accurate evaluation. The format of a sample in TabIS is $(T, Q, R, O)$ where $Q$ is a question, $R, O$ are correct and wrong options. In TabIS,  $T$ and $R$ are the same as the annotation in the TTG task,  $O$ is a wrong description of the table that we generate, and $Q$ is a question about the table that can be answered by $R$. So, the task of TabIS is that, given $T$ and $Q$, choose an option from $\{R, O\}$ as the answer.


TabIS contains three subsets: basic table information seeking (B-TIS), TIS requiring structure understanding (SU-TIS), and TIS from multiple tables (M-TIS). In the following, we will first introduce how to generate options, and then introduce these subsets respectively.

\begin{figure*}[t]
\centering
\includegraphics[width=0.98\textwidth]{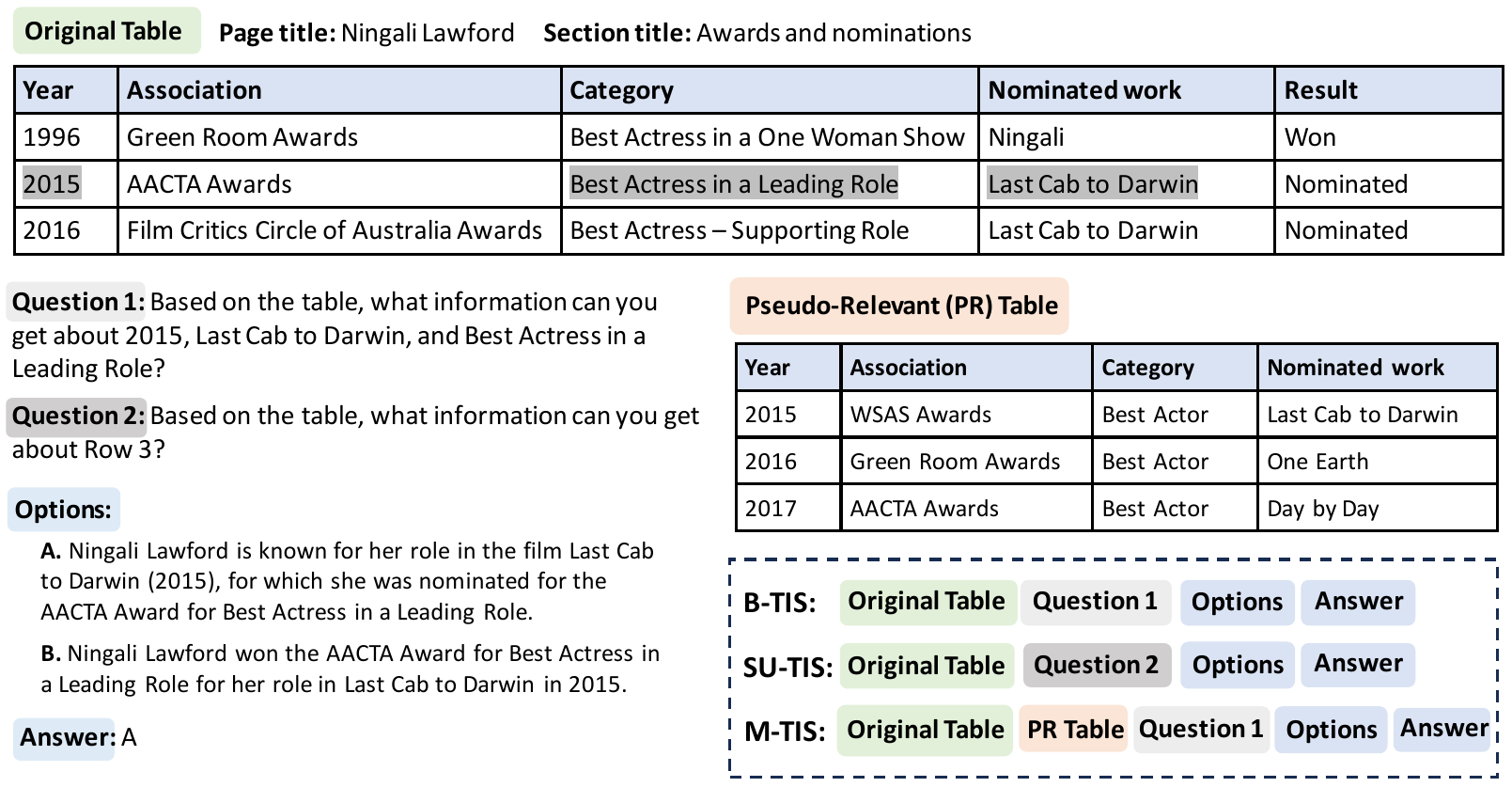}
\caption{Simplified Examples of B-TIS subset, SU-TIS subset, and M-TIS subset. For each B-TIS sample, we generate one SU-TIS sample and one M-TIS sample with some modifications. }
\label{fig:dataset_example}
\end{figure*}

\subsection{Option Generation Method}\label{sec:sis_option_generation}

The option generation has three steps:
\begin{enumerate}[leftmargin=0.5cm, itemsep=0pt, parsep=0pt, topsep=0pt, partopsep=0pt]
    \item For each TTG sample, we generate one challenging candidate option, expecting that the option is unfaithful to the table but is similar to the golden reference.
    \item We perform adversarial filtering~\citep{evaleval} to divide all instances into easy and hard categories. Specifically, we use three different LLMs on two different presentation orders of the options ($R, O$ and $O,R$) to obtain six predicted labels. The instances in which the majority of labels are wrong are hard instances and others are easy instances. (Discussed in Appendix \ref{app_diff_div}.)
    \item For hard instances, we conduct manual checking and modification on generated options to ensure correctness.
\end{enumerate}

In step 1, three strategies to generate options are proposed:

\noindent\textbf{Modify-Input (MI).}\quad   We directly prompt GPT-4 to first modify the highlighted cells 
$C$ slightly, resulting in a modified set $C'$, and subsequently perform the TTG task using $C'$ to produce an unfaithful statement $O$ referring to $R$. The generated $O$ usually has a similar syntactic structure as $R$ but substitutes some entities.

\noindent\textbf{Modify-Output (MO).}\quad   We directly prompt GPT-4 to refer to the golden reference $R$ and make up a new statement that contains highlighted cells $C$, but is not faithful to the table fact. 

\noindent\textbf{Exam-Judge (EJ).}\quad Given the table $T$ and a set of cells $C$, we first instruct a weak LLM agent to describe the cells in natural language, yielding multiple candidate responses $\{O'_1, O'_2, \dots \}$. Subsequently, a more advanced LLM agent\footnote{We use GPT-3.5-turbo-16k and GPT-4 as the weak and strong LLM agent, respectively.} is employed to identify responses that are unfaithful to the table. Among these unfaithful candidates, the one that is most literally similar to the golden reference $R$ is selected as the wrong option (detailed in Appendix \ref{app_choose_opt}). The underlying idea is to automatically obtain incorrect responses from relatively weak agents, thereby producing strong false options that are diverse and deceptive. In the experiments, we find this method is good at generating difficult instances.

In step 3, for hard instances, we instruct annotators to check if the generated option is faithful to the table. If it is faithful, then it needs to be revised to an unfaithful description while ensuring the altered options are convincingly deceptive.

Finally, each instance can be categorized into four classes, \textbf{MI, MO, EJ}, and \textbf{HA} (Human-Annotation, i.e. modified in step 3) according to how its $O$ is generated. We put more details of the option generation pipeline in Appendix \ref{app_opt_gen}. 

\subsection{B-TIS Subset}\label{sec:b_tis}

B-TIS mimics situations where the LLM agent is tasked with offering clear statements to users who inquire about specific real-world entities, such as celebrities and sports events, based on a table. This method could markedly diminish the necessity for users to sift through massive table data, which is crucial in areas such as sport summary writing and financial data analysis. We show an example in Figure \ref{fig:dataset_example}.

We apply the aforementioned option generation pipeline to generate the B-TIS dataset using two public TTG datasets: (1) \textbf{ToTTo}~\citep{totto} is an open-domain English table-to-text dataset with over 120,000 examples. The tables in ToTTo are all semi-structured HTML tables from Wikipedia pages and the reference sentences are mainly descriptive statements over the table fact.
(2) \textbf{HiTab}~\citep{hitab} is a cross-domain hierarchical table dataset with over 10,000 samples, constructed from a wealth of statistical reports. It contains hierarchical tables and accompanied descriptive sentences collected from StatCan and NSF.  Compared to ToTTo, HiTab poses a greater challenge to table information seeking since the tables are with hierarchies and the sentences may involve numerical reasoning (e.g. comparison and simple computation).

\subsection{SU-TIS Subset}

In LLM-based chat systems like ChatGPT~\citep{chatgpt}, a straightforward way for users to direct the LLM agent to a specific area of a table is by indicating positions (e.g., "row 3").
This requires LLMs to understand table structures. We mimic this scenario by introducing the TIS dataset that emphasizes structural understanding (SU-TIS). For each instance $(T, Q, R, O)$ in B-TIS, we modify question $Q$ by replacing the selected cells with the minimum set of rows or columns covering them, as illustrated in Figure \ref{fig:dataset_example}.

\subsection{M-TIS Subset}


In real-world scenarios, LLM agents may be presented with additional context that, while superficially related to the golden table (the table that contains the answer), could be misleading and detrimentally affect their information seeking capabilities~\citep{lost_in_the_middle}. 
This situation frequently arises in retrieval-augmented LLM systems oriented to documents, where in response to a query, the systems may retrieve several tables that are relevant to the query but not golden.

To mimic this scenario, we investigate the effects of adding one pseudo-relevant table, which appears relevant to the main table but does not provide useful information to answer the question. We show an example in Figure \ref{fig:dataset_example}. For each instance in B-TIS, we add another table $T'$ to the tuple $(T, Q, R, O)$, resulting in $(\{T, T'\}, Q, R, O)$. $T'$ is generated by prompting GPT-4 to create one table mirroring the structure and headers of the golden table, yet contains varied data entries. Refer to Appendix \ref{app_mtis} for more details.

\subsection{Dataset Statistics and Quality Assessment}

\renewcommand{\arraystretch}{1.4} 
\begin{table}[t]\centering
\setlength\tabcolsep{7pt}
\fontsize{9}{9}\selectfont
\begin{threeparttable}
\begin{tabular}{ll|cc}
\toprule
\multicolumn{2}{l|}{\textbf{Dataset}}     & \textbf{\# Train} & \textbf{\# Test} \\ 
\hline
\multirow{2}{*}{B-TIS}     & ToTTo & 20,244    & 1,283    \\
                         & HiTab & 6,943     & 1,254    \\ \hline
\multirow{2}{*}{SU-TIS} & ToTTo & 20,054    & 1,267    \\
                         & HiTab & 6,864     & 1,215    \\ \hline
\multirow{2}{*}{M-TIS} & ToTTo & 0        & 1,217    \\
                         & HiTab & 0        & 1,139    \\ \hline
\multicolumn{2}{l|}{Total}       & 54,105    & 7,375    \\
\bottomrule
\end{tabular}
\end{threeparttable}
\caption{Data statistics of TabIS. \label{tab:statistic}}
\end{table}

\renewcommand{\arraystretch}{1.4} 
\begin{table}[t]\centering
\setlength\tabcolsep{4pt}
\fontsize{9}{9}\selectfont
\begin{threeparttable}
\begin{tabular}{l|ccc|ccc}
\toprule   & \textbf{ToTTo} & \textbf{Ratio} & \textbf{Acc.} & \textbf{HiTab} & \textbf{Ratio} & \textbf{Acc.} \\
\hline
MI & 433                & 33.7\%    &  93.5\%     & 345            & 27.5\%     &  90.5\%     \\
MO & 495                & 38.6\%   & 95.8\%      & 366                & 29.2\%   &  97.2\%      \\
EJ & 267                & 20.8\%  & 91.7\%       & 438                & 34.9\%    & 89.2\%     \\
HA & 88                 & 6.9\%   & 100.0\%       & 105                & 8.4\%  &100.0\% \\
\bottomrule
\end{tabular}
\end{threeparttable}
\caption{Statistics of option generation strategies used in B-TIS datasets. Acc. denotes data accuracy assessed by experts. \label{tab:stat_opt}}
\end{table}

Table \ref{tab:statistic} illustrates the data statistics of the datasets used in our experiments.
We show the statistics of the strategies used for generating options in Table~\ref{tab:stat_opt}. We engage 10 sophisticated annotators to meticulously review and revise the hard instances in the test set (step 3 in Section \ref{sec:sis_option_generation}). Out of 410 reviewed samples, the options for 193 samples are manually adjusted. We employ two experts to assess the data accuracy on 200 samples each from ToTTo and HiTab. The overall accuracy of ToTTo and HiTab is 94.1\% and 92.5\%, respectively, demonstrating the high quality of the proposed TabIS. 


%% file: experiment2.tex
\section{Experiments on TabIS}

Based on the curated TabIS benchmark, we evaluate the table information seeking capabilities of 12 representative LLMs.

\renewcommand{\arraystretch}{1.3} 
\begin{table*}[th]\centering
\setlength\tabcolsep{7pt}
\fontsize{9}{9}\selectfont
\begin{threeparttable}
\begin{tabular}{l|cccccc|c}
\toprule
\multicolumn{1}{l|}{\multirow{2}{*}{\textbf{Model}}} & \multicolumn{2}{c}{\textbf{B-TIS}} & \multicolumn{2}{c}{\textbf{SU-TIS}} & \multicolumn{2}{c|}{\textbf{M-TIS}} & \multirow{2}{*}{\textbf{Avg.}} \\
\multicolumn{1}{l|}{}                                & \textbf{ToTTo}  & \textbf{HiTab} & \textbf{ToTTo}    & \textbf{HiTab}   & \textbf{ToTTo}    & \textbf{HiTab}    &                                \\ \hline
\textit{proprietary model}                          &                 &                &                   &                  &                   &                  &                                \\
Gemini-pro                                          & 85.6            & 66.6           & 81.3              & 65.1             & 79.4              & 64.8             & 73.8                           \\
GPT-3.5-turbo-instruct                              & 75.1            & 68.3           & 70.8              & 65.3             & 74.5              & 66.8             & 70.1                           \\
GPT-3.5-turbo-1106                                  & 72.1            & 57.5           & 66.8              & 50.4             & 66.7              & 53.0             & 61.1                           \\
GPT-3.5-turbo-16k                                   & 76.7            & 61.2           & 73.3              & 59.2             & 73.4              & 59.2             & 67.2                           \\
GPT-4-turbo-1106                                    & \textbf{91.2}   & \textbf{82.4}  & \textbf{90.0}     & \textbf{81.7}    & \textbf{89.7}     & \textbf{80.4}    & \textbf{85.9}                  \\
\midrule
\textit{open-source model}                          &                 &                &                   &                  &                   &                  &                                \\
Llama2-7b-chat                                      & 53.6            & 47.8           & 53.1              & 48.8             & 52.3              & 48.6             & 50.7                           \\
TableLlama-7b                                       & 54.3            & 47.7           & 54.1              & 47.8             & 54.1              & 47.9             & 51.0                           \\
Mistral-7b-instruct-v0.2                            & 73.2            & 56.9           & 69.9              & 53.5             & 68.8              & 57.1             & 63.2                           \\
Llama2-13b-chat                                     & 63.3            & 53.4           & 57.9              & 50.5             & 60.5              & 54.4             & 56.7                           \\
Mixtral-8*7b-instruct                               & 80.6            & 65.6           & 80.8              & \textbf{62.7}    & 76.2              & 57.9             & 70.6                           \\
Llama2-70b-chat                                     & 70.0            & 56.9           & 67.8              & 54.3             & 67.4              & 54.7             & 61.9                           \\
Tulu2-70b-DPO                                       & \textbf{85.7}   & \textbf{68.2}  & \textbf{81.9}     & 61.9             & \textbf{82.9}     & \textbf{64.0}    & \textbf{74.1}    \\
\bottomrule
\end{tabular}
\end{threeparttable}
\caption{Main results on TabIS. Random-guess achieves a 50\% accuracy. \label{tab:tis_main}}

\end{table*}

\subsection{Experimental Settings}\label{sec:exp_setting}

\noindent\textbf{Problem settings.}\quad  We evaluate LLMs in a table-based QA setting, where a linearized markdown table is presented in the context, and LLMs are required to answer a question given the context. All the questions are constructed into the single-choice form with two options, as detailed in Section \ref{tabis_bench}. We use a \textbf{one-shot example}\footnote{We find that more examples would often surpass the 4,096 token limit commonly used by open-source models.} to familiarize the model with the task description and answering format, similar to previous work~\citep{tulu}. Refer to Appendix \ref{app_mtis} for more details.

We evaluate both proprietary and open-source LLMs. To enhance reproducibility, we set the temperature as 0 for proprietary models, and utilize the maximum probability of the first token as A or B to determine the outputs of open-source models.

\noindent\textbf{Proprietary models.}\quad 
We adopt three representative models:  \textbf{GPT-3.5}~\citep{chatgpt},  \textbf{GPT-4}~\citep{gpt4} and \textbf{Gemini-pro}~\citep{gemini-pro}.  GPTs\footnote{For GPTs, we investigate \textit{GPT-3.5-turbo-1106} and \textit{GPT-4-turbo-1106} for more consistent evaluation. We also report results on \textit{GPT-3.5-turbo-instruct} and \textit{GPT-3.5-turbo-16k}, since we find their performance varies greatly.} is a series of popular and capable LLM systems developed by OpenAI. Recent studies~\citep{numerical_reasoning,gpt4table,invest_ttg} have shown the great potential of these models on table-related tasks. Gemini-pro\footnote{Gemini-pro is currently accessible via the \href{https://makersuite.google.com/app}{Gemini API}.} is Google's most capable LLM which operates seamlessly across various modalities.

\noindent\textbf{Open-source models.}\quad 
Using proprietary LLM APIs as agents presents many challenges such as high costs and privacy concerns \citep{evaleval}. Therefore, we evaluate several popular open-source models: (1) \textbf{Llama2-chat}~\citep{llama2} ranging from 7b to 70b parameters; (2) \textbf{Mistral-7b-instruct-v0.2}~\citep{mistral-7b} and \textbf{Mixtral-8x7b-instruct}~\citep{mixtral-8*7b}, an instruction-tuned sparse mixture of experts language model; (3) \textbf{TableLlama-7b}~\citep{tablellama}, instruction-tuned from Llama2-7b, the first large generalist models for tables; and (4) \textbf{Tulu2-70b-DPO}~\citep{tulu2}, finetuned from Llama2-70b, the first 70b model aligned with DPO~\citep{dpo}. These models represent the highest-quality LLMs of different architectures and alignment strategies available to the community. 

\subsection{Main Results on TabIS}

We show the results of various models on the test set of TabIS in Table \ref{tab:tis_main}. Refer to Appendix \ref{app_tis_result} for more details.


\noindent\textbf{Overall Performance.}\quad As shown in the ``Avg.'' column in Table~\ref{tab:tis_main}, both proprietary models and open-source models perform poorly in TabIS. Proprietary models are generally superior to open-source models, with the highest average accuracy recorded at 85.9 by GPT-4-turbo, compared to 74.1 by Tulu2-70b-DPO. Gemini-pro outperforms GPT-3.5s but falls short of GPT-4-turbo.
Regarding open-source models, a trend is observed where larger models within the same series generally outperform their smaller counterparts. For instance, Llama2-chat models with 7b, 13b, and 70b parameters achieve average accuracies of 50.7, 56.7, and 61.9, respectively.
However, this trend does not hold across different model series, where a larger model size does not guarantee superior performance. For example, the 7b version of Mistral-instruct even surpasses the 70b Llama2-chat model by 1.3 points. This observation raises an important question about the impact of pre-training and alignment strategies on the TIS capabilities of LLMs which may be an interesting research topic.

\noindent\textbf{Performance on TabIS Subsets.}\quad The middle columns in Table~\ref{tab:tis_main} show that all models generally perform better in B-TIS compared to SU-TIS and M-TIS, indicating SU-TIS and M-TIS are more challenging. SU-TIS, which only provides the location of highlighted cells as hints, are inherently more difficult than B-TIS. However, models can refer to the cells contained in options to look back at the table to verify each option, therefore the performance drop is not dramatic.
M-TIS introduces an extra table that is only seemingly relevant, potentially confusing the judgement of LLMs.
In comparisons between datasets, all models show better performance on ToTTo than on HiTab, with improvements ranging from 5.8 to 19.0 points. This discrepancy is likely due to ToTTo predominantly featuring standard tables without merged cells, whereas HiTab includes tables with complex hierarchies, which pose greater challenges for table comprehension. 

\begin{figure}[t]
\centering
\includegraphics[width=0.48\textwidth]{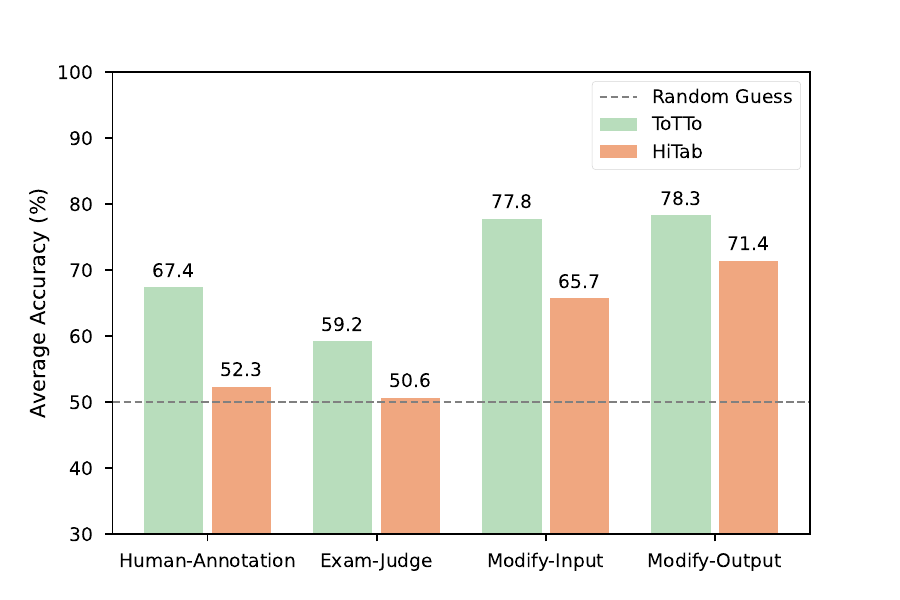}
\caption{Model performance in different option generation strategies. Averaged over 12 LLMs.}
\label{fig:opt_strategy}
\end{figure}

\noindent\textbf{Comparing option generation strategies.}\quad
As illustrated in Figure \ref{fig:opt_strategy}, models exhibit the lowest performance with options generated via Exam-Judge, with average scores of only 59.2 and 50.6 for ToTTo and HiTab, respectively. This indicates that Exam-Judge is capable of producing options that are even more challenging for LLMs than those annotated by humans. Modify-Input and Modify-Output also present significant hurdles for LLMs, with scores ranging from 65.7 to 78.3 points on average. For options generated by humans, while they are tough enough, they also lead to high expenses. Our option generation pipeline leverages the advanced instruction-following capabilities of potent LLMs, effectively balancing cost-efficiency with scalability.

\section{Potential Challenges}

In this section, we conduct in-depth analysis to investigate the LLMs' limitations and potential challenges behind the two complex subsets: SU-TIS (Section \ref{sec:tsu}) and M-TIS (Section \ref{sec:pseudo_relevant}). We further show error analysis on hard samples in Section \ref{sec:case_study}.

\subsection{Table Structure Understanding}\label{sec:tsu}

We further investigate the table structure understanding (TSU) capabilities of LLMs, shedding light on future research on the SU-TIS sub-task.

TSU refers to the ability to perceive the two-dimensional layout inherent in tables, such as the positioning of cells, rows, and columns, to access desired content based on the location within the table space. 
TSU is highly important to our SU-TIS, which involves locating a specific region of the table.
While this may seem intuitive to humans, it can be quite challenging for LLMs, especially because tables are fed to these models in a serialized format, such as markdown or HTML.
To investigate the TSU capabilities of LLMs, we design six basic lookup tasks, such as "What is the content of cells in row 3/column 3?" and "What is the content of cells within the same row as the cell 'Harry Potter'?" We employ predefined templates to generate test samples from semi-structured HTML tables, transforming them into a single-choice format with two options. Each sample includes one in-context example, similar to TabIS. Refer to Appendix \ref{app_tsu_data} for more details.

\begin{figure}[t]
\centering
\includegraphics[width=0.48\textwidth]{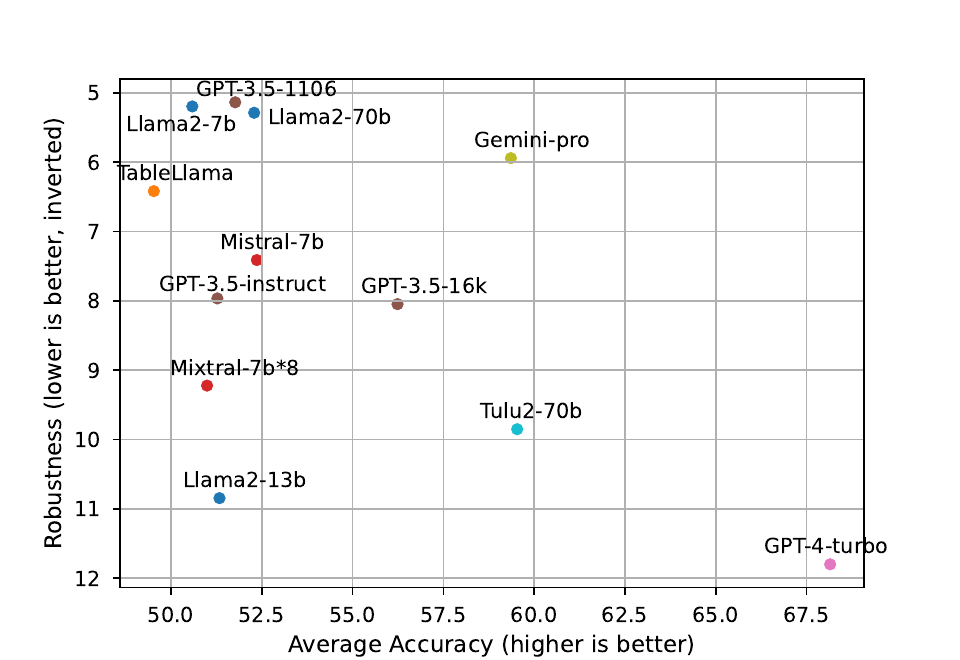}
\caption{Averaged accuracy and TSU variation score for 12 models, tested and averaged on 6 TSU tasks. Model names are simplified for illustration.}
\label{fig:pis_pos_2d}
\end{figure}

Once humans understand the table structure and the task description, their TSU performance ideally remains excellent and consistent regardless of target locations. However, we find that LLMs work in a totally different manner.
Specifically, we report the average accuracy on six tasks and the variation score towards target positions in Figure \ref{fig:pis_pos_2d}. The variation score for a TSU task is defined as the standard deviation in accuracy across different target locations. 
Notably, most LLMs achieve near-random performance (50) on TSU tasks. The strongest LLM, GPT-4-turbo, exhibits the lowest stability. No LLMs stand out in both performance and stability. Refer to Appendix \ref{app_tsu_exp} for additional analysis.

This highlights a common challenge of table structure understanding: \textbf{LLMs exhibit poor performance on TSU tasks and the accuracy varies greatly across different positions.} In real-world scenarios of SU-TIS, there are no options for a user query. LLMs can only locate the target region based on the positional information (e.g. row 3). The TIS performance would be largely affected by models' TSU capabilities. We will also release the six TSU datasets to facilitate future research.

\subsection{Robustness against Pseudo-Relevant Tables}\label{sec:pseudo_relevant}

\begin{figure}[t]
\centering
\includegraphics[width=0.48\textwidth]{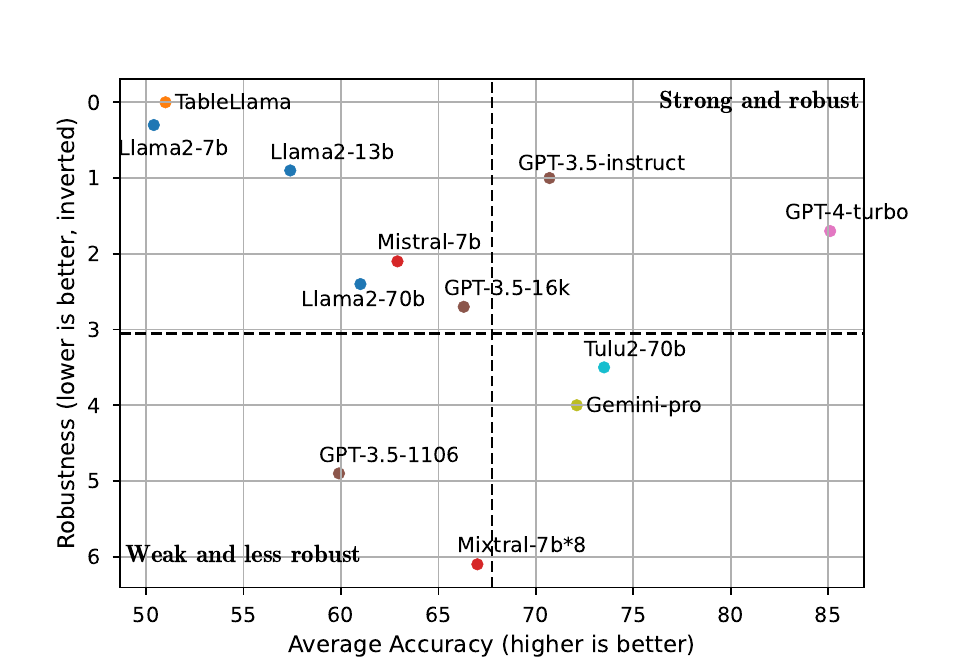}
\caption{TIS Robustness against pseudo-relevant tables and averaged accuracy for 12 models, tested and averaged on ToTTo and HiTab. Model names are simplified for illustration.}
\label{fig:rag_scatter}
\end{figure}

Based on M-TIS, we further investigate the TIS robustness of various models against pseudo-relevant tables. Specifically, to quantify a model's robustness, we measure the deviation between the accuracy without and with the pseudo-relevant table, averaged on ToTTo and HiTab. 
The results are shown in Figure \ref{fig:rag_scatter}.
Notably, GPT-3.5-instruct and GPT-4-turbo emerge as both effective and robust. However, the two strongest open-source models, Tulu-70b and Mixtral-7b*8, exhibit the lowest robustness levels. Besides, within the same model series, larger models achieve better accuracy scores but worse robustness scores. This phenomenon can be observed in Llama2 series (7b, 13b, 70b) and Mistral series (Mistral-7b, Mixtral-8*7b).
M-TIS indicates \textbf{great challenges of LLMs in balancing between TIS performance and robustness against pseudo-relevant tables, especially for open-source models}. This finding calls for future research on open-source models to improve TIS robustness against pseudo-relevant tables.

\subsection{Error Analysis on Hard Samples}\label{sec:case_study}

To explore why LLMs fall short on TabIS, we conduct further analyses based on the hard set of B-TIS. Specifically, we sample 50 instances from the hard B-TIS set and ask an expert to analyze the reasons why these questions are difficult to answer. Finally, we categorize the main types of difficulties into four categories, as shown in Table \ref{tab:error_analysis}. We find that current LLMs still make mistakes in distinguishing subtle details and are more prone to commit errors on  options that appear more concrete but contain errors (R1, R3). Additionally, table information seeking often requires numerical reasoning (R2) and common sense knowledge (R4), areas where current LLMs are not yet proficient.

\begin{table*}[ht]
  \centering
  \fontsize{10}{10}\selectfont
  \begin{tabular}{l|l|c|c}
    \toprule
     & \textbf{Reason Type} & \textbf{Ratio} & \textbf{Example} \\
    \midrule
    R1 & The wrong option seems more concrete but contains errors in the details. & 45.6\% & Figure \ref{fig:case_study_r1_totto} \\
    R2 & Involving numerical reasoning (comparison/simple calculations). & 21.2\% & Figure \ref{fig:case_study_r2_hitab} \\
    R3 & The wrong option is constructed by replacing one cell text with its nearby cell. & 15.2\% & Figure \ref{fig:case_study_r3_totto} \\
    R4 & Requiring an overall understanding combining common sense and table semantics. & 18.2\% & Figure \ref{fig:case_study_r4_totto} \\
    
    \bottomrule
  \end{tabular}
  \caption{Four main reason types of hard samples. We show the ratio of each type and one example for illustration. }
  \label{tab:error_analysis}
\end{table*}

\section{Improving Table Information Seeking}

In this section, we explore how supervised finetuning enhances table information seeking using weakly-supervised datasets.

We first utilize our proposed data generation pipeline\footnote{Considering high cost of accessing GPT-4 API, we use GPT-3.5-turbo-16k instead.} (Section \ref{tabis_bench}) to construct weakly-supervised B-TIS and SU-TIS training datasets without manual checking. The statistics of the training dataset are shown in Table \ref{fig:dataset_example}.  We fully finetune \textit{Llama2-13b-chat} on this training set for 2 epochs to obtain \textbf{TISLlama}. We evaluate TISLlama on TabIS\footnote{Note that training on the weakly-supervised datasets may introduce the spurious correlation between the model-generated options and the wrong options. Thus we only evaluate on human-annotated samples for fair comparision.}. Refer to Appendix \ref{app_train} for more training details.

Table \ref{tab:sft_held_in} demonstrates that TISLlama outperforms both the base model Llama2-13b-chat and the leading open-source model Tulu2-70b-DPO, with margins of 17.7 and 5.4 points, respectively.
These results demonstrate the effectiveness of TIS-oriented supervised finetuning. However, its performance does not yet match that of GPT-4-turbo, which has not undergone specialized fine-tuning. This discrepancy highlights the significant challenge TabIS presents to large language models, underscoring the need for further research in this area.


\renewcommand{\arraystretch}{1.3} 
\begin{table}[H]\centering
\setlength\tabcolsep{3pt}
\fontsize{9}{9}\selectfont
\begin{threeparttable}
\begin{tabular}{lcccc}
\toprule
\textbf{Model}   & \textbf{B-TIS} & \textbf{SU-TIS} & \textbf{M-TIS} & \textbf{Avg.} \\
\midrule
Llama2-13b-chat  & 56.8         & 53.3             & 56.5             & 55.5          \\
Llama2-70b-chat  & 58.2         & 58.1             & 58.5             & 58.3          \\
Tulu2-70b-DPO $~\clubsuit$    & 69.7         & 69.1             & 64.7             & 67.8          \\
GPT-4-turbo-1106 $\spadesuit$  & 81.2         & 77.4             & 79.1             & 79.2          \\
\midrule
TISLlama (ours)   & 73.3      & 73.7        & 72.7        & 73.2 \\
\bottomrule
\end{tabular}
\end{threeparttable}
\caption{Evaluation of TISLlama on TabIS-HA, averaged on ToTTo and HiTab. $~\clubsuit$ and $\spadesuit$ denote the best open-source and proprietary model in our evaluation. \label{tab:sft_held_in}}
\end{table}

%% file: related_work.tex
\section{Related Work}

\subsection{Table-to-Text generation}
Table-to-Text generation (TTG) aims at generating natural language statements that faithfully describe the information contained in the provided table region. 
Given its broad applications like biographical data analysis~\citep{app_biograph} and sports game summary generation~\citep{app_sport}, TTG has been studied extensively in recent years~\citep{lattice,openrt} with the introduction of several valuable datasets~\citep{totto,hitab,nli_acc}.
Previous studies mainly focus on finetuning pre-trained language models on a task-specific dataset~\citep{lattice}, which are often specialized  and lack generalizability. 
Large language models (LLMs) have recently demonstrated remarkable performance on TTG tasks~\citep{distill_table_reason,invest_ttg}. However, these evaluations mainly rely on surface-level metrics, such as BLEU~\citep{bleu} and ROUGE~\citep{rouge}, which may result in unreliable evaluation when the syntactic style of LLMs' response diverges from the golden reference~\citep{parent}. In this paper, we employ the TTG tasks as a test bench for evaluating table information seeking of LLMs. To ensure a reliable assessment, we construct single-choice questions based on two high-quality TTG datasets, ToTTo~\citep{totto} and HiTab~\citep{hitab}.

\subsection{Evaluating Table Information Seeking capabilities of LLMs}

Short-form table QA datasets, such as WikiSQL~\citep{wikisql} and WikiTQ~\citep{wikitablequestions}, contain queries seeking for information, such as "who is the manufacturer for the order year 1998?". However, these datasets focus on relational tables and evaluate the ability to comprehend table schemas and transform natural language queries into SQL queries, while our work emphasizes grasping the information conveyed by the complex table contents. Besides, Free-form table QA datasets such as FeTaQA~\citep{fetaqa} relies on text-similarity-based metrics leads to unreliable evaluations, a problem exacerbated in the era of LLMs. 

Recently, \citet{gpt4table} presents a benchmark designed to measure the structural comprehension of large language models (LLMs) through the comparison of various input approaches. However, it solely examines the capabilities of the most advanced LLM, GPT-4. Their findings suggest that GPT-4 possesses a fundamental understanding of table structures, yet there's a noticeable absence of comprehensive evaluations across a wider range of LLMs and an examination of TSU consistency.
\citet{invest_ttg} investigates the potential of applying LLMs in real-world table information seeking scenarios, showcasing their effectiveness in producing faithful statements. Nevertheless, their analysis is significantly influenced by unreliable evaluation metrics.

To the best of our knowledge, we are the first to release a large-scale, comprehensive, reliable benchmark for evaluating TIS capabilities.

%% file: conclusion.tex
\section{Conclusion}

This paper introduces TabIS, a new benchmark designed to evaluate the table information seeking (TIS) abilities of large language models (LLMs). TabIS is comprised of three typical TIS scenarios and employs a single-choice question format to ensure reliable evaluation. Extensive experiments on 12 representative LLMs have shown that TabIS presents a significant challenge for current LLMs, with GPT-4-turbo showing only marginal satisfaction. Further analysis points out two main issues: firstly, LLMs perform almost randomly on basic tasks involving comprehension of table structures; secondly, they face difficulties in improving performance and maintaining robustness against pseudo-relevant tables, which could lead to sub-optimal results in real-world TIS tasks. These observations underscore the current limitations and potential challenages in TIS, calling for further exploration and advancement in this area.

%% file: limitation.tex
\section{Limitations}

In this paper, the benchmark adopts the form of single-choice questions, which ensures the reliability of the evaluation but may deviate from practical applications. TabIS is design with only two options, which may not sufficiently challenge LLMs, particularly when GPT-4 shows high baseline accuracy. The templates used for generating TIS questions are relatively simplistic; richer and more diverse questions would enhance the quality of the benchmark. 
We use GPT-4 to modify prompts and generate pseudo-relevant tables, which may introduce bias, potentially favoring GPT-series models due to their inherent familiarity with the dataset construction.

Given that the tables originate from Wikipedia, there may be concerns regarding data contamination; LLMs might still perform well without the context provided by the tables. We show a discussion on Appendix \ref{app_data_cont}.
Besides, we mainly discuss some limitations and potential challenges of LLMs when handling table information seeking tasks, but do not explore how to address these issues or the reasons behind their observations.


\section*{Acknowledgements}

This work has been supported by the National Natural Science Foundation of China (No. 62076231, 62206265), and the China Postdoctoral Science Foundation (No. 2021M703271). We thank all the anonymous reviewers for their valuable and constructive comments.

%% file: appendix.tex
\newpage

\appendix



\section{Option Generation Details}\label{app_opt_gen}
\begin{figure*}[t]
\centering
\includegraphics[width=0.95\textwidth]{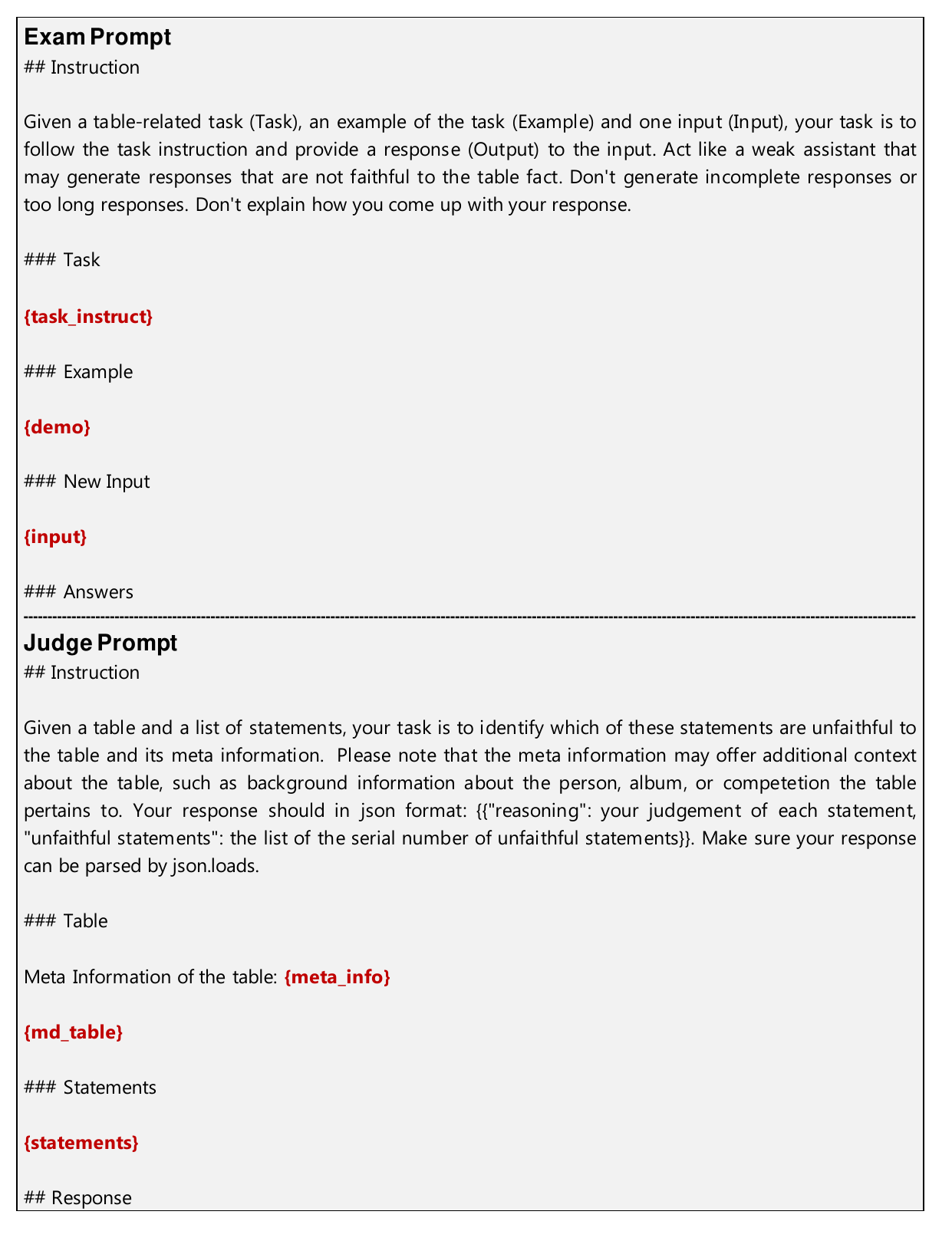}
\caption{Prompt of Exam-Judge.}
\label{fig:exam-judge}
\end{figure*}

\begin{figure*}[t]
\centering
\includegraphics[width=0.95\textwidth]{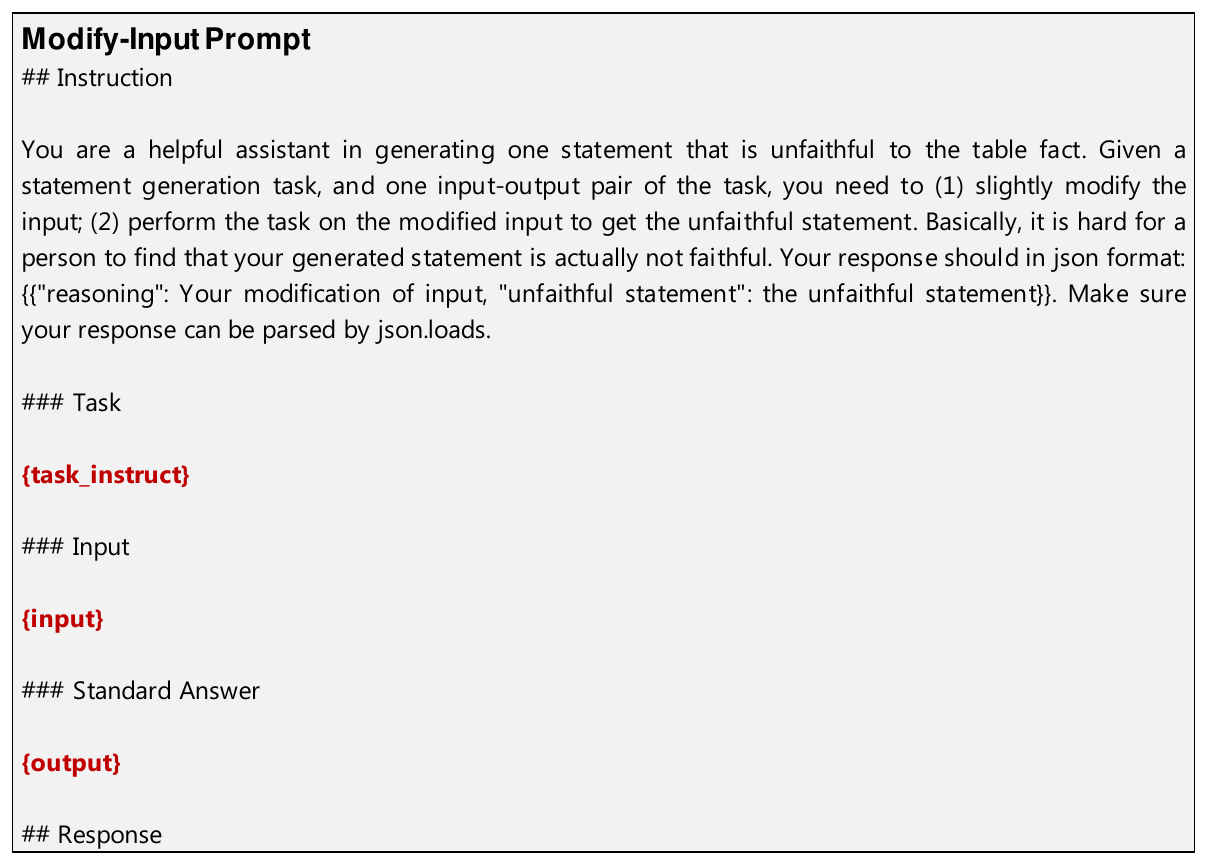}
\caption{Prompt of Modify-Input.}
\label{fig:mod-input}
\end{figure*}

\begin{figure*}[t]
\centering
\includegraphics[width=0.95\textwidth]{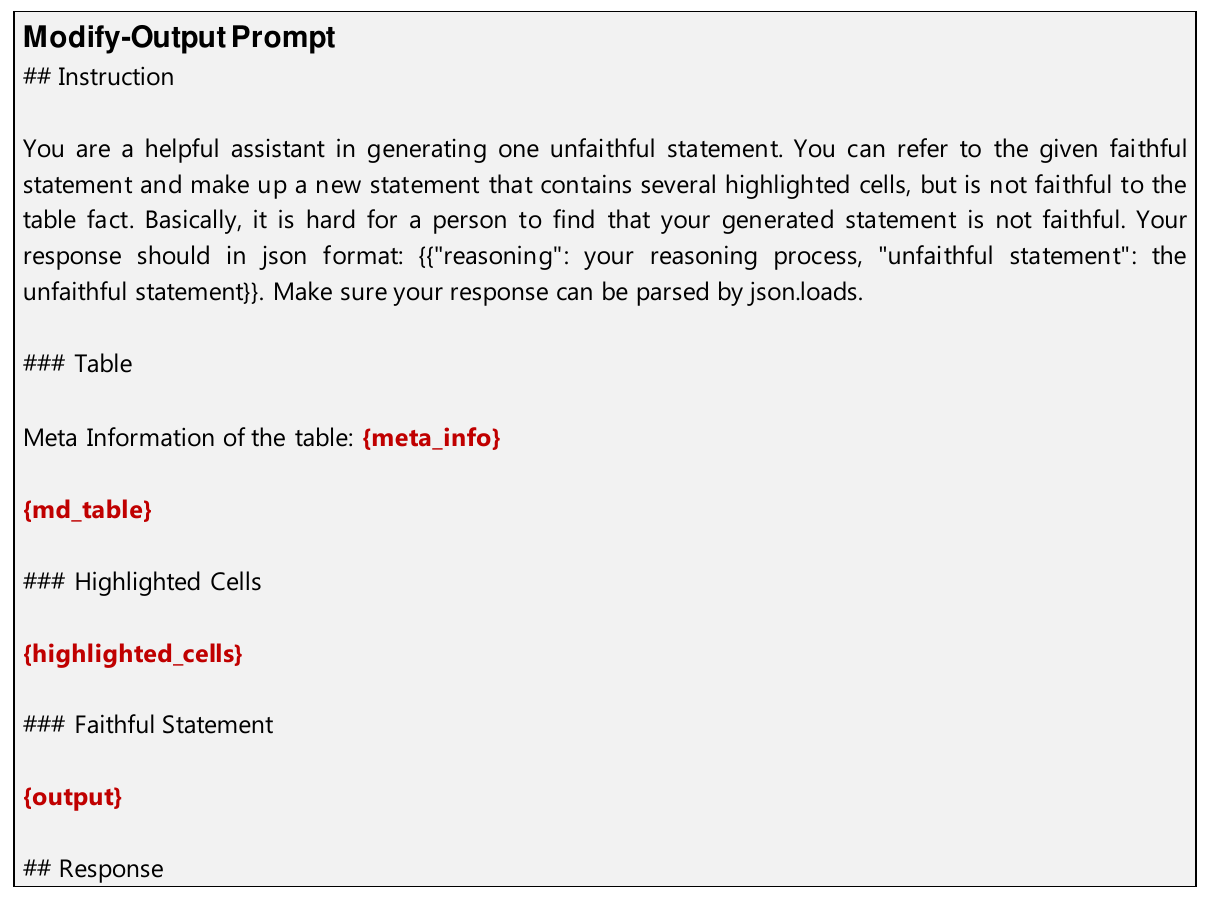}
\caption{Prompt of Modify-Output.}
\label{fig:mod-output}
\end{figure*}

For each TTG sample, we apply one of the three option generation strategies to generate one candidate option. Considering cost and quality, we use GPT-3.5-turbo-16k, GPT-3.5-turbo-instruct, and GPT-3.5-turbo-1106 to perform the adversarial filtering. 
We show the prompt of Exam-Judge, Modify-Input, and Modify-Output in Figure \ref{fig:exam-judge}, Figure \ref{fig:mod-input}, and Figure \ref{fig:mod-output}, respectively.

\subsection{Discussion on Difficulty Division}\label{app_diff_div}
In our experiments, we use adversarial filtering to determine the difficulty of samples. The classification achieved through this process is indeed accurate, with the easy samples being less challenging than the more complex ones. This assertion is supported by two pieces of evidence: 
\begin{itemize}
    \item \textbf{Model performance}. It's observed that models typically fare much better on easy samples than on hard ones. To demonstrate, we present the performance of various models on the B-TIS-ToTTo in Table \ref{tab:diff_div}. Both proprietary and open-source models exhibit superior performance on easier samples.
    \item \textbf{Expert assessment effort}. On average, experts can assess easy samples at twice the rate (32 per hour) of hard samples (14 per hour), highlighting a significant difference in the complexity and time required for evaluation.
\end{itemize}

\begin{table}[htbp]
  \centering
  \fontsize{10}{10}\selectfont
  \begin{tabular}{l|ccc}
    \toprule
    \textbf{Model} & \textbf{Easy} & \textbf{Hard} & \textbf{$|\Delta|$}\\
    \midrule
    Llama2-7b-chat & 56.9 & 49.7 & 7.2 \\
    Llama2-13b-chat & 70.9 & 54.2 & 16.7 \\
    Llama2-70b-chat & 81.4 & 56.3 & 25.1 \\
    GPT-3.5-turbo-1106 & 92.8 & 57.1 & 35.7 \\
    GPT-4-1106-preview & 99.7 & 81.0 & 18.7 \\
    \bottomrule
  \end{tabular}
  \caption{Model performance on the easy subset and hard subset of B-TIS-ToTTo.}
  \label{tab:diff_div}
\end{table}

\subsection{Details on Choosing Option Candidates}\label{app_choose_opt}
We simply determine the similarity by comparing the sentence lengths (measured in characters) of the golden reference and the option candidates. The premise is that an option with a length similar to that of the golden reference is likely to be more misleading. Based on our observations, these option candidates tend to be semantically similar but differently phrased. Therefore, we just choose the option that is closer in length to the standard answer.

\section{Benchmark Details}\label{app_mtis}

As described in Section \ref{sec:exp_setting}, each test sample within our benchmark is accompanied by one in-context example. We only keep samples that contain fewer than 4000 tokens when processed by LlamaTokenizer\footnote{Note that different LLMs may use different tokenizers. Besides, the tokenizers of proprietory models are unaccessible. Therefore, we choose the tokenizer of Llama2, one of the most popular open-source models.}~\citep{llama2}, in adherence to the common 4096 token limit imposed by most open-source models.

\noindent\textbf{SU-TIS.}\quad Given a set of row-column coordinates $C=\{(r_i, c_i)\}_{i=1}^{|C|}$ representing highlighted cells in a B-TIS sample, we construct the SU-TIS question using the smaller set between $\{r_i \mid (r_i, c_i) \in C\}$ and $\{c_i \mid (r_i, c_i) \in C\}$.

\begin{figure*}[t]
\centering
\includegraphics[width=0.95\textwidth]{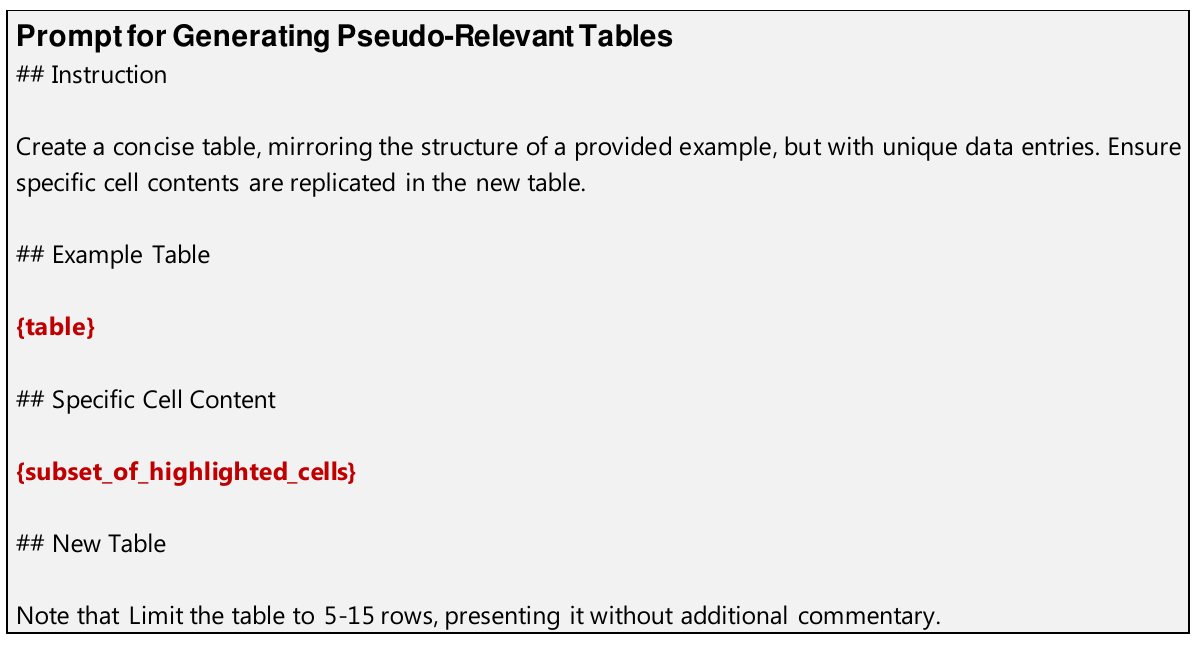}
\caption{Prompt for generating pseudo-relevant tables.}
\label{fig:gen-prt}
\end{figure*}

\noindent\textbf{M-TIS.}\quad We use GPT-4-turbo-1106 to generate one pseudo-relevant table for each B-TIS sample. We show the prompt for generating pseudo-relevant tables in Figure \ref{fig:gen-prt}. We randomly placed the noisy table $T'$ either before or after the golden table $T$.  In earlier trials, we attempted to retrieve similar tables from the ToTTo training set to serve as pseudo-relevant tables. However, the retrieved tables often had weak relevance, likely because the dataset's various tables are sourced from different HTML documents. Considering the challenges in acquiring noisy tables from realistic retrieval scenarios (e.g., within the same document), we opted to generate noise tables with high similarity to the golden table using LLMs. 

\section{Exploring Table Structure Understanding}\label{app_tsu}
In this section, we first introduce the construction of TSU dataset, then we show our additional experiments on TSU. 

\subsection{TSU Dataset Construction}\label{app_tsu_data}
Understanding the structure of a table is essential for navigating among data arranged in a tabular format, interpreting the relations among data points, and understanding the table semantics. It requires to perceive the two-dimensional spatial layout inherent in tables, such as the positioning of cells, rows, and columns, to access desired content based on the location within the table space. 

To examine the table structure understanding capabilities of LLMs, we propose six probing tasks: positional cell lookup (PCL), relative cell lookup (RCL), positional row lookup (PRL), relative row lookup (RRL), positional column lookup (PLL), relative column lookup (RLL). These tasks require LLMs to acquire certain surface-level table components (cell, row and column) based on relative or absolute position information.

We generate samples for each task by applying predefined templates on high-quality tables. All question templates are shown in Table \ref{tab:pis_template}. We collect tables from four public datasets: WikiSQL~\citep{wikisql}, WikiTableQuestions~\citep{wikitablequestions}, HybridQA~\citep{hybridqa} and FeTaQA~\citep{fetaqa}. These tables are all semi-structured HTML tables collected from Wikipedia, spaning a wide array of topics such as sports and geography. After deduplicating these tables, we obtain a total of 49,561 high-quality tables. For the test set, we randomly sample 1\% tables and generate one sample per table for each task.

For each sample, the options are generated by randomly sampling cells, rows and columns in proximity to the golden answer, employing a gaussian distribution $\mathcal{N}(\mathbf{p}, 1)$, where $\mathbf{p}$ denotes the position of the golden answer.

\begin{figure}[H]
\centering
\includegraphics[width=0.48\textwidth]{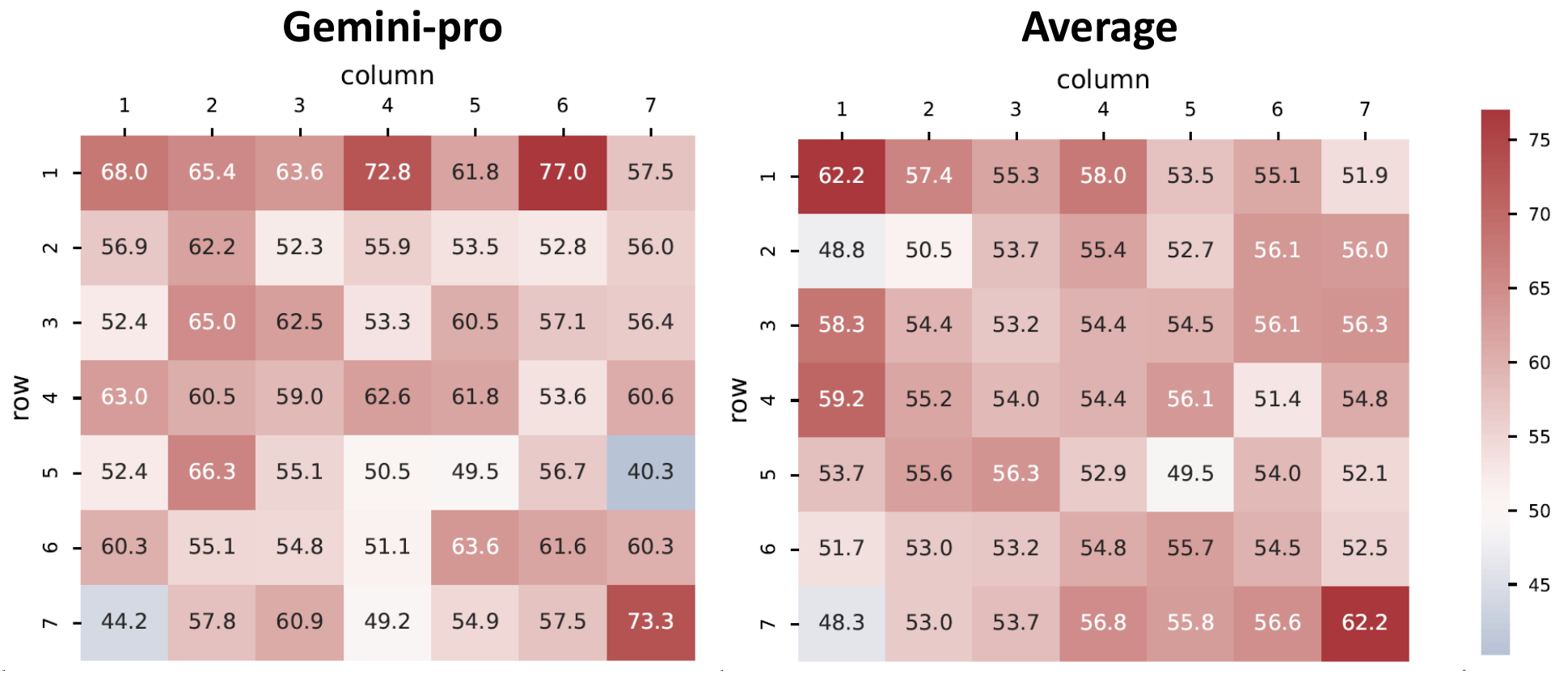}
\caption{RCL performance with respect to target cell positions. We show a concrete example of Gemini-pro (left) and the averaged results of 12 models (right).}
\label{fig:evid_pos_heatmap}
\end{figure}

\subsection{Additional Experiments}\label{app_tsu_exp}

We show the TSU performance of various models on Table \ref{tab:tsu_main}.

\noindent\textbf{TSU Performance.}\quad
Unexpectedly, despite TSU being straightforward for humans, all LLMs demonstrate subpar performance. The best performance of proprietary models and open-source models only achieve 66.1 (GPT-4) and 57.6 points (Tulu2-70b), respectively, while most models achieve near-random performance (50). 
Models do not consistently excel across all types of TSU tasks. Notably, the GPT series (GPT-4 and GPT-3.5) tend to perform better in column-oriented tasks (PLL, RLL) relative to other tasks, whereas the Llama2 series (Llama2-7b, 13b, 70b) show greater proficiency in cell-oriented tasks (PCL, RCL). This variation in performance could be attributed to the fact that models within the same series likely undergo similar pre-training and alignment processes, resulting in comparable inductive biases.


\noindent\textbf{Case Study on Variations across different positions.}\quad
As presented in Section \ref{sec:tsu}, LLMs exhibit fluctuating performance across different positions. We further show case studies of RCL in Figure \ref{fig:evid_pos_heatmap}. For example, Gemini-pro exhibits large variance in different positions,with a disparity exceeding 30 points between its highest and lowest accuracy. Similar patterns are noted in other LLMs. On average, the data indicates that LLMs perform more effectively at the beginning (row 1, column 1) and ending (row 7, column 7) of tables. This pattern is likely influenced by the serialization of tables into one-dimension strings, rendering the middle part of the table more challenging to locate accurately. 

\noindent\textbf{Effect of Cell Contents on TSU.}\quad
Logically, carrying out TSU tasks should be independent of the particular content within table cells, since this doesn't necessitate grasping the table's underlying semantics. Thus, the performance across tables with varying content should be consistent. To test this, we altered the cell contents in our TSU test set's real tables to random numbers (ranging from 1 to 8 digits) and random letters (also 1 to 8 characters in length), creating two new synthetic test sets named "letter" and "number."
 
\begin{figure}[H]
\centering
\includegraphics[width=0.48\textwidth]{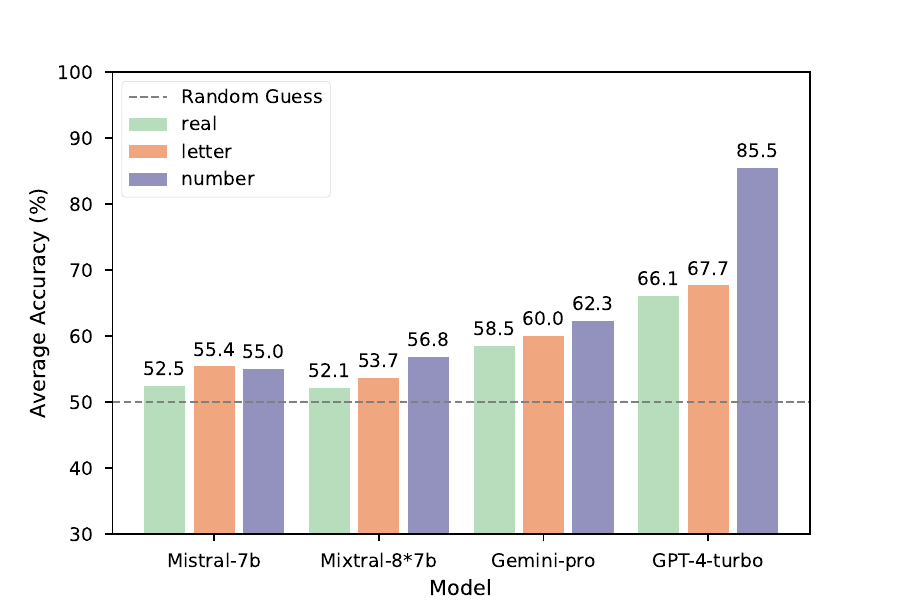}
\caption{Accuracy on tables of different content, averaged on 6 TSU tasks.}
\label{fig:pis_content}
\end{figure}

However, we observe \textbf{significant variation in performance across different table contents}. As shown in Figure \ref{fig:pis_content}, the performance disparity between the test sets ranges from approximately 2.9 to 19.4 points. Intriguingly, GPT-4 shows markedly improved performance on the "number" set. This may be attributed to the activation of GPT-4's numerical processing capabilities, which are particularly relevant for TSU tasks (e.g. counting rows). This observation warrants further investigation in future studies.

\subsection{Summary}
These findings show that current LLMs exhibit weak performance in TSU tasks, both in terms of overall performance and consistency across varying positions and cell contents. 



\section{TabIS Main Results}\label{app_tis_result}
We show the main results of B-TIS, SU-TIS, and M-TIS in Table \ref{tab:tis_all}, Table \ref{tab:tis_tsu_all}, and Table \ref{tab:tis_prt_all}, respectively. We also report the accuracy on each option generation strategies.

We present some results of statistical significance testing in Table \ref{tab:significance}. Specifically, we assess whether the test accuracy significantly surpasses random guessing (50\% accuracy). All p-values are below 0.05, indicating that the results are statistically significant.

\begin{table}[htbp]
  \centering
  \fontsize{10}{10}\selectfont
  \begin{tabular}{l|cc}
    \toprule
    \textbf{Model} & \textbf{Acc.} & \textbf{p-value} \\
    \midrule
    Llama2-7b-chat & 53.6 & $5.09 \times 10^{-3}$ \\
    Llama2-13b-chat & 63.3 & $6.58 \times 10^{-22}$ \\
    Llama2-70b-chat & 70.0 & $6.55 \times 10^{-48}$ \\
    GPT-3.5-turbo-1106 & 72.1 & $1.86 \times 10^{-58}$ \\
    GPT-4-1106-preview & 91.2 & $3.13 \times 10^{-222}$ \\
    \bottomrule
  \end{tabular}
  \caption{p-values of significance tests for different Models on B-TIS (ToTTo). Typically, p-values less than 0.05 are considered significant.}
  \label{tab:significance}
\end{table}

\section{Training Details}\label{app_train}

We fully fine-tune the model \textit{Llama2-13b-chat}\footnote{https://huggingface.co/meta-llama/Llama-2-13b-chat-hf} with LlaMA-Factory~\citep{llama-factory}. We use a learning rate of 2e-5. We train the model on 8 A800 and use a linear scheduler with a 5\% warm-up period for 2 epochs. To efficiently train the model, we employ DeepSpeed training with ZeRO-3 stage~\citep{zero}. For both training and inference, we set the input length as 4096.

\section{Case Study}\label{app_train}

\begin{figure*}
\centering
\includegraphics[width=0.95\textwidth]{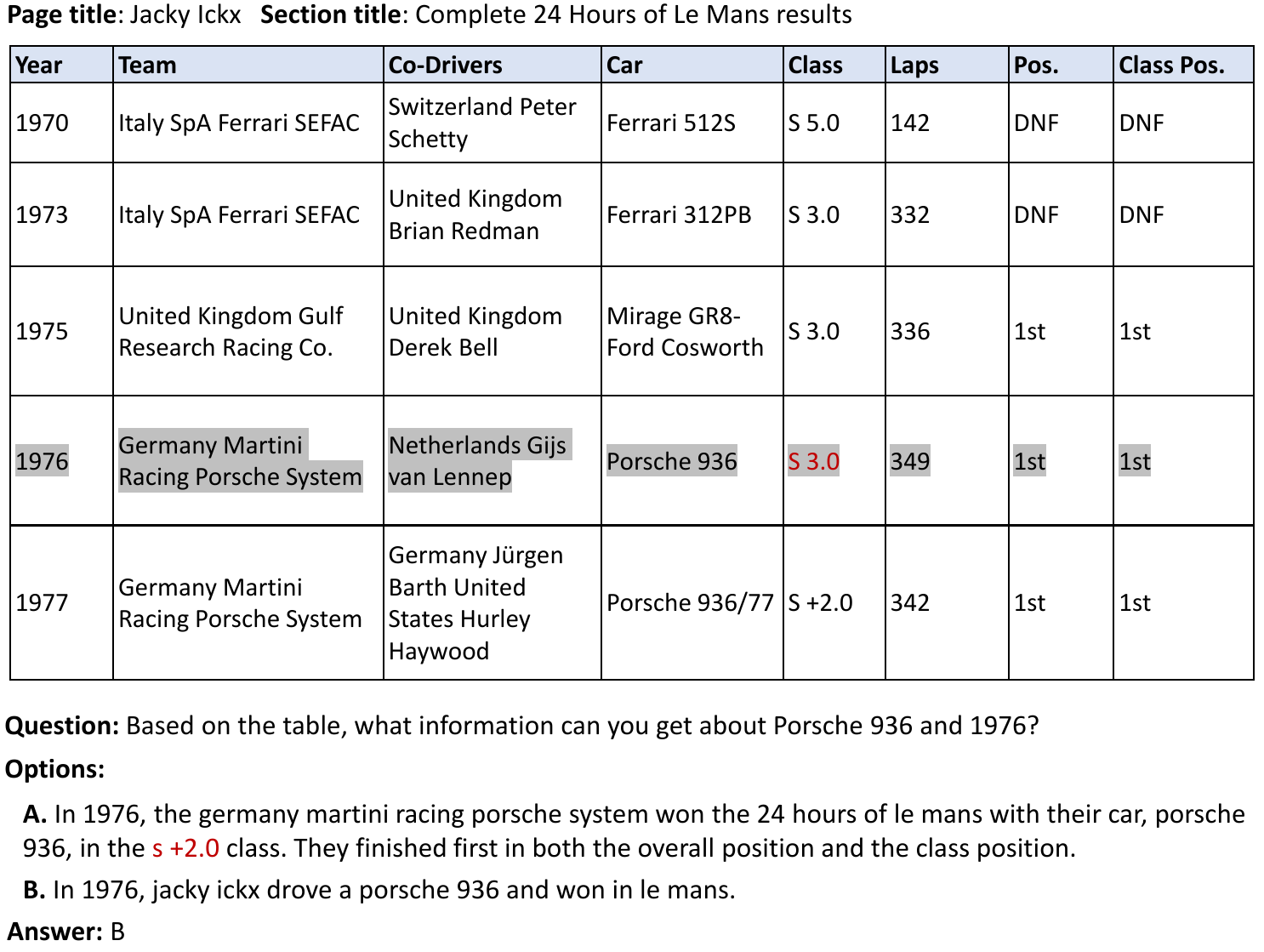}
\caption{Example (Simplified) of R1 from ToTTo. Relevant cells are highlighted. Option A seems more concrete but contains incorrect details. The class should be "S 3.0" rather than "S +2.0". }
\label{fig:case_study_r1_totto}
\end{figure*}

\begin{figure*} 
\centering
\includegraphics[width=0.95\textwidth]{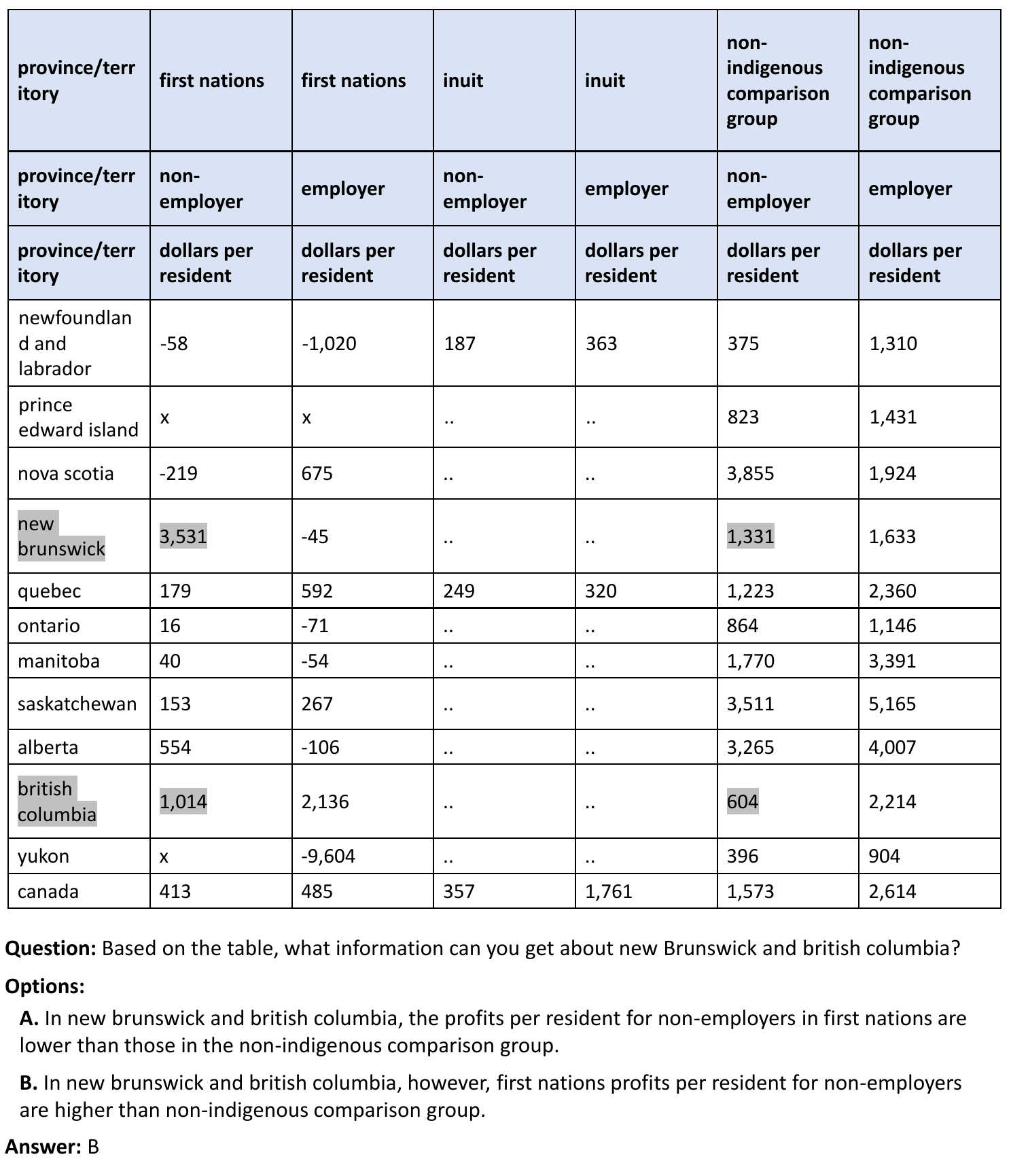}
\caption{Example (Simplified) of R2 from HiTab. Relevant cells are highlighted. Both options require comparing numbers in the table.}
\label{fig:case_study_r2_hitab}
\end{figure*}

\begin{figure*} 
\centering
\includegraphics[width=0.90\textwidth]{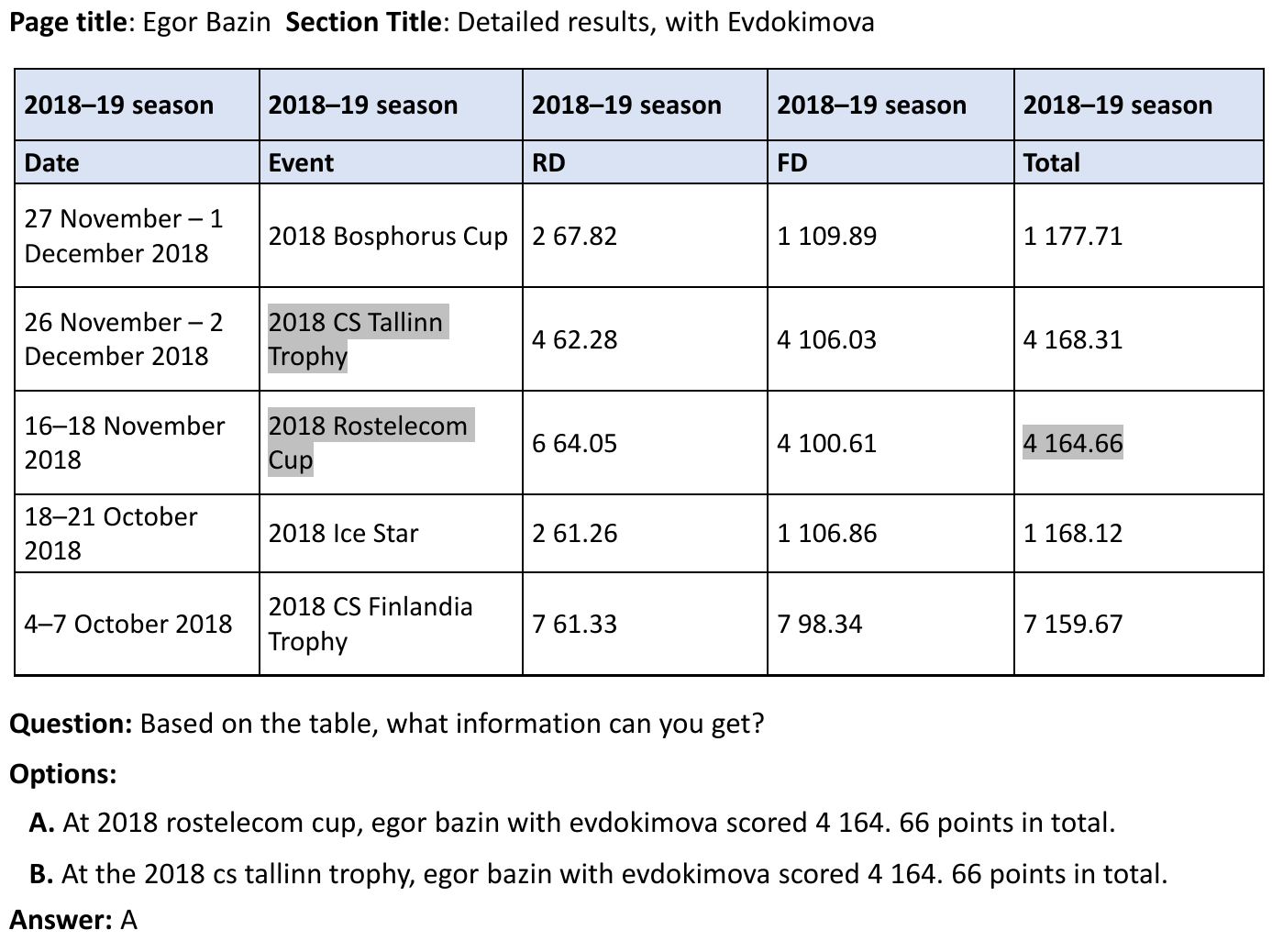}
\caption{Example (Simplified) of R3 from ToTTo. Relevant cells are highlighted. Option B replaces "2018 Rostelecom Cup" with the upper cell "2018 CS Tallinn
Trophy" in the table. }
\label{fig:case_study_r3_totto}
\end{figure*}

\begin{figure*} 
\centering
\includegraphics[width=0.90\textwidth]{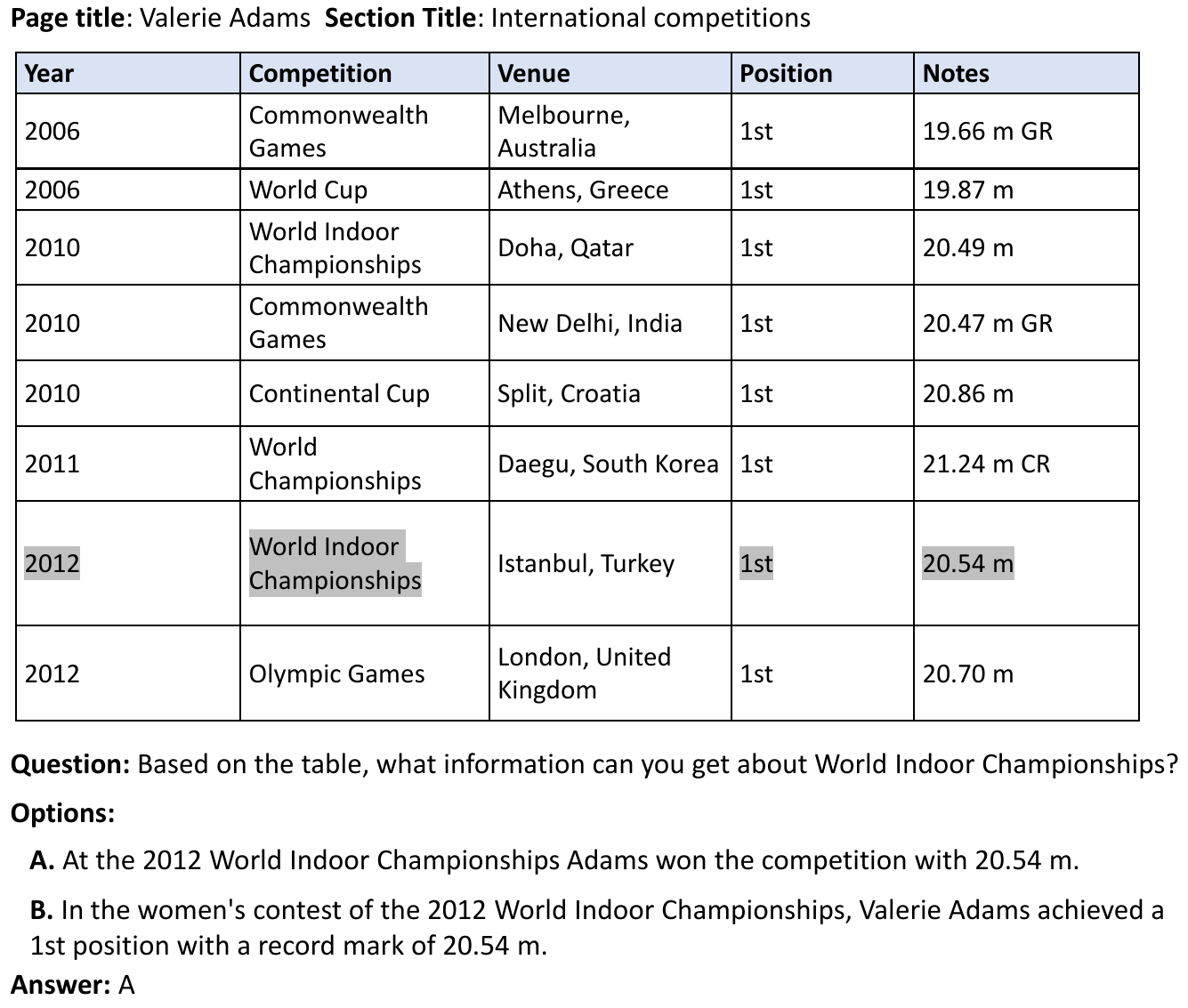}
\caption{Example (Simplified) of R4 from ToTTo. Relevant cells are highlighted. Option B requires understanding the meaning of data in the "Notes" column and knowing that GR and CR indicate a record-breaking performance.}
\label{fig:case_study_r4_totto}
\end{figure*}

We show examples of four reason types in Figure \ref{fig:case_study_r1_totto}, Figure \ref{fig:case_study_r2_hitab}, Figure \ref{fig:case_study_r3_totto}, and Figure \ref{fig:case_study_r4_totto}.

\section{Discussion on Data Contamination}\label{app_data_cont}
Given that the tables originate from Wikipedia, there may be concerns regarding data contamination; LLMs might still perform well without the context provided by the tables. This issue can be intensive especially when the correct option is factually correct with external world knowledge while the incorrect option is not.
To investigate this problem, we conduct experiments to assess how models' parametric knowledge influence their performance on choosing the correct option. We remove the tables from the test questions, allowing the models to answer based solely on their world knowledge. The results on 500 BIS samples are shown in Table \ref{tab:data_cont}. "w. table" denotes the model performance on the original dataset. "w/o. table" denotes performance after table removal. $\Delta$ denotes the performance gain purely from the table understanding.

\begin{table}[htbp]
  \centering
  \fontsize{10}{10}\selectfont
  \begin{tabular}{l|ccc}
    \toprule
    \textbf{Model} & \textbf{w. table} & \textbf{w/o. table} & \textbf{$\Delta$}\\
    \midrule
    \textit{ToTTo} & & & \\
    Llama2-7b-chat & 53.6 & 52.2 & 1.4 \\
    Llama2-13b-chat & 63.3 & 56.0 & 7.3 \\
    Llama2-70b-chat & 70.0 & 55.4 & 15.6 \\
    GPT-3.5-turbo-1106 & 75.1 & 56.8 & 18.3 \\
    \hline
    \textit{HiTab} & & & \\
    Llama2-7b-chat & 47.8 & 47.2 & 0.6 \\
    Llama2-13b-chat & 53.4 & 46.0 & 7.4 \\
    Llama2-70b-chat & 56.9 & 48.0 & 8.9 \\
    GPT-3.5-turbo-1106 & 68.3 & 52.4 & 15.9 \\
    \bottomrule
  \end{tabular}
  \caption{Model performance on 500 B-TIS samples with/without the table.}
  \label{tab:data_cont}
\end{table}

After removing the table, all models exhibit similar performance, which is nearly at random, with ToTTo slightly higher than HiTab. The ranking of different models based on the delta values is consistent with the ranking using the original dataset. Despite of data contamination, the dataset can authentically reflect the table information seeking capabilities of models.

\renewcommand{\arraystretch}{1.7} 
\centering
\fontsize{7}{7}\selectfont
\begin{table*}[]
\begin{tabular}{p{4.5cm}|p{10.5cm}}
\hline
\textbf{Task (T)}                                      & \textbf{Question Template (Q)}                                                                                                                       \\ \hline
\multirow{1}{*}{Positional Cell   Lookup}   
                                            & Q: What is the content of the  cell located at row \{row\} and column \{col\}?                                                                                          \\\hline
\multirow{1}{*}{Positional Row   Lookup}  & Q: What are the contents of the cells in row \{row\}?                                                                                                                  \\\hline
\multirow{1}{*}{Positional Column   Lookup}  & Q: What are the contents of   the cells in column \{col\}?                                                                                                               \\\hline
\multirow{4}{*}{Relative Cell   Lookup}   & Q1: The anchor cell is  \{anchor\} in row \{row\} and column \{col\}. What is the content of the first cell   below the anchor cell within the same column?              \\
                                            & Q2: The anchor cell is  \{anchor\} in row \{row\} and column \{col\}. What is the content of the first cell   above the anchor cell within the same column?              \\
                                            & Q3: The anchor cell is  \{anchor\} in row \{row\} and column \{col\}. What is the content of the first cell   left to the anchor cell within the same row?                \\
                                            & Q4: The anchor cell is  \{anchor\} in row \{row\} and column \{col\}. What is the content of the first cell   right to the anchor cell within the same row?              \\\hline
\multirow{3}{*}{Relative Row Lookup}     & Q1: The anchor cell is \{anchor\} in row \{row\} and column \{col\}. What are the contents of the cells within the   same row as the anchor cell?                         \\
                                            & Q2: The anchor cell is \{anchor\} in row {row} and column {col}. What are the contents of the first row above the anchor cell?                         \\
                                            & Q3: The anchor cell is \{anchor\} in row \{row\} and column \{col\}. What are the contents of the first row below the anchor cell?                         \\\hline
\multirow{3}{*}{Relative Column Lookup}   & Q1: The anchor cell is \{anchor\}   in row \{row\} and column \{col\}. What are the contents of the cells within the   same column as the anchor cell? \\
                                            & Q2: The anchor cell is \{anchor\} in row \{row\} and column \{col\}. What are the contents of the first column left to the anchor cell?  \\ 
                                            & Q3: The anchor cell is \{anchor\} in row \{row\} and column \{col\}. What are the contents of the first column right to the anchor cell? \\ \hline
\end{tabular}
\caption{TSU Tasks (T) and corresponding Question Templates (Q). Placeholders \{row\}, \{col\}, and \{anchor\} represent the row number, column number, and the content of the anchor cell, respectively.}\label{tab:pis_template}
\end{table*}

\renewcommand{\arraystretch}{1.4} 
\begin{table*}[t]
\centering
\fontsize{10}{10}\selectfont
\begin{threeparttable}
\begin{tabular}{l|cccccc|c}
\toprule
\textbf{Model}                      & \textbf{PCL} & \textbf{PRL} & \textbf{PLL} & \textbf{RCL} & \textbf{RRL} & \textbf{RLL} & \textbf{Avg.} \\ \hline
\textit{proprietary model} &               &               &               &               &               &               &               \\
Gemini-pro                 & 50.7          & \textbf{59.9} & 46.2          & 51.3          & 70.3          & 72.9          & 58.5          \\
GPT-3.5-turbo-16k          & \textbf{55.1} & 53.1          & 61.5          & 54.9          & 55.7          & 54.7          & 55.8          \\
GPT-3.5-turbo-instruct     & 47.5          & 46.1          & 56.9          & 40.0          & 63.7          & 56.8          & 51.8          \\
GPT-3.5-turbo-1106         & 50.4          & 50.8          & 53.6          & 49.8          & 53.2          & 49.9          & 51.3          \\
GPT-4-turbo-1106           & 50.2          & 38.3          & \textbf{82.4} & \textbf{72.7} & \textbf{74.7} & \textbf{78.3} & \textbf{66.1} \\
\hline
\textit{open-source model} &               &               &               &               &               &               &               \\
Llama2-7b-chat             & \textbf{53.3} & 47.8          & 50.0          & 55.7          & 47.8          & 50.1          & 50.8          \\
TableLlama-7b              & 49.2          & \textbf{53.7} & 53.6          & 55.1          & 54.4          & 54.3          & 53.4          \\
Mistral-7b-instruct-v0.2   & 49.0          & 45.9          & 52.9          & 58.0          & 56.7          & 52.6          & 52.5          \\
Llama2-13b-chat            & 51.6          & 51.8          & 51.9          & 57.8          & 53.2          & 52.2          & 53.1          \\
Mixtral-8*7b-instruct      & 47.1          & 48.0          & \textbf{55.9} & 52.7          & 57.1          & 52.2          & 52.1          \\
Llama2-70b-chat            & 51.6          & 48.2          & 47.5          & 56.5          & 51.3          & 47.8          & 50.5          \\
Tulu2-70b-DPO              & 50.6          & 48.6          & 54.8          & \textbf{67.1} & \textbf{70.0} & \textbf{54.5} & \textbf{57.6} \\
\bottomrule
\end{tabular}
\caption{Main results (accuracy) of various models across TSU tasks. Random-guess achieves a 50\% accuracy. \label{tab:tsu_main}}
\end{threeparttable}
\end{table*}

\renewcommand{\arraystretch}{1.4} 
\begin{table*}[]\centering
\fontsize{10}{10}\selectfont
\begin{threeparttable}
\begin{tabular}{@{}l|ccccc|ccccc@{}}
\toprule
\multicolumn{1}{c|}{\multirow{2}{*}{\textbf{Model}}} & \multicolumn{5}{c|}{\textbf{ToTTo}}                                    & \multicolumn{5}{c}{\textbf{HiTab}}                                    \\
\multicolumn{1}{c|}{}                                & EJ & MI & MO & HA & \textbf{Avg.} & EJ & MI & MO & HA & \textbf{Avg.} \\ \midrule
\textit{proprietary model}                           &               &               &               &               &               &               &               &               &               &               \\
gemini-pro                                           & 70.2          & 93.3          & 87.9          & 76.9          & 85.6          & 53.1          & 67.6          & 79.1          & 67.4          & 66.6          \\
GPT-3.5-turbo-instruct                               & 60.7          & 81.8          & 80.6          & 55.7          & 75.1          & 62.3          & 71.9          & 78.4          & 45.7          & 68.3          \\
GPT-3.5-turbo-1106                                   & 56.9          & 77.8          & 76.6          & 64.8          & 72.1          & 42.5          & 64.6          & 71.3          & 48.6          & 57.5          \\
GPT-3.5-turbo-16k                                    & 58.4          & 84.5          & 82.8          & 59.1          & 76.7          & 48.4          & 67.5          & 75.4          & 43.8          & 61.2          \\
GPT-4-turbo-1106                                     & \textbf{79.8} & \textbf{93.5} & \textbf{96.4} & \textbf{85.2} & \textbf{91.2} & \textbf{73.5} & \textbf{85.2} & \textbf{91.8} & \textbf{77.1} & \textbf{82.4} \\
\midrule
\textit{open-source model}                                 &               &               &               &               &               &               &               &               &               &               \\
Llama2-7b-chat                                       & 54.3          & 52.4          & 53.1          & 60.2          & 53.6          & 44.3          & 54.8          & 47.8          & 39.1          & 47.8          \\
TableLlama-7b                                        & 53.2          & 54.7          & 53.9          & 58.0          & 54.3          & 43.8          & 53.3          & 48.9          & 41.0          & 47.7          \\
Mistral-7b-instruct-v0.2                             & 52.8          & 77.4          & 81.0          & 70.5          & 73.2          & 40.9          & 63.5          & 72.4          & 47.6          & 56.9          \\
Llama2-13b-chat                                      & 52.4          & 66.7          & 66.7          & 60.2          & 63.3          & 45.0          & 52.2          & 64.8          & 53.3          & 53.4          \\
Mixtral-8*7b-instruct                                & 55.8          & 88.7          & 88.1          & 73.9          & 80.6          & 51.6          & \textbf{75.1} & 77.1          & 52.4          & 65.6          \\
Llama2-70b-chat                                      & 52.1          & 70.9          & 79.6          & 65.9          & 70.0          & 46.8          & 60.0          & 68.0          & 50.5          & 56.9          \\
Tulu2-70b-DPO                                        & \textbf{64.4} & \textbf{91.7} & \textbf{93.1} & \textbf{78.4} & \textbf{85.7} & \textbf{55.5} & 72.5          & \textbf{81.4} & \textbf{61.0} & \textbf{68.2} \\
\bottomrule
\end{tabular}
\caption{B-TIS Main results on ToTTo and HiTab. We also report accuracy across different option generation strategies: EJ (Exam-Judge), MI (Modify-Input), MO (Modify-Output), HA (Human-Annotation). \label{tab:tis_all}}
\end{threeparttable}
\end{table*}

\renewcommand{\arraystretch}{1.4} 
\begin{table*}[]\centering
\fontsize{10}{10}\selectfont
\begin{threeparttable}
\begin{tabular}{@{}l|ccccc|ccccc@{}}
\toprule
\multicolumn{1}{l|}{\multirow{2}{*}{\textbf{Model}}} & \multicolumn{5}{c|}{\textbf{ToTTo}}   & \multicolumn{5}{c}{\textbf{HiTab}}                                            \\
\multicolumn{1}{l|}{}                                & \textbf{EJ}   & \textbf{MI}   & \textbf{MO}   & \textbf{HA}   & \textbf{Avg.} & \textbf{EJ}   & \textbf{MI}   & \textbf{MO}   & \textbf{HA}   & \textbf{Avg.} \\
\midrule
\textit{proprietary model}                          &               &               &               &               &               &               &               &               &               &               \\
gemini-pro                                          & 72.2          & 81.8          & 87.6          & 64.7          & 81.3          & 52.6          & 71.4          & 75.7          & 54.0          & 65.1          \\
GPT-3.5-turbo-instruct                              & 55.9          & 75.7          & 78.5          & 48.9          & 70.8          & 57.4          & 69.9          & 75.4          & 48.5          & 65.3          \\
GPT-3.5-turbo-1106                                  & 49.8          & 72.5          & 72.5          & 58.0          & 66.8          & 34.9          & 60.2          & 62.0          & 42.7          & 50.4          \\
GPT-3.5-turbo-16k                                   & 56.7          & 81.3          & 78.5          & 55.7          & 73.3          & 43.3          & 69.6          & 62.0          & 42.7          & 59.2          \\
GPT-4                                               & \textbf{77.6} & \textbf{91.7} & \textbf{96.5} & \textbf{83.0} & \textbf{90.0} & \textbf{71.0} & \textbf{85.2} & \textbf{94.1} & \textbf{71.8} & \textbf{81.7} \\
\midrule
\textit{open-source model}                                &               &               &               &               &               &               &               &               &               &               \\
Llama2-chat-7b                                      & 52.1          & 52.1          & 53.3          & 60.2          & 53.1          & 45.9          & 55.4          & 48.4          & 40.8          & 48.8          \\
TableLlama-7b                                       & 53.2          & 52.8          & 54.8          & 59.1          & 54.1          & 44.5          & 51.8          & 49.6          & 42.7          & 47.8          \\
Mistral-7b-instruct-v0.2                            & 48.3          & 74.3          & 78.3          & 65.9          & 69.9          & 34.4          & 63.0          & 71.1          & 41.8          & 53.5          \\
Llama2-chat-13b                                     & 51.3          & 59.7          & 61.0          & 52.3          & 57.9          & 42.9          & 49.1          & 60.1          & 54.4          & 50.5          \\
Mixtral-8*7b-instruct                               & 56.3          & \textbf{88.0} & 88.2          & 78.4          & 80.8          & \textbf{49.0} & \textbf{71.1} & 73.9          & 54.4          & \textbf{62.7} \\
Llama2-chat-70b                                     & 51.0          & 68.8          & 75.4          & 71.6          & 67.8          & 44.7          & 60.5          & 62.9          & 44.7          & 54.3          \\
Tulu2-70b-DPO      & \textbf{63.1} & 85.9          & \textbf{88.6} & \textbf{81.8} & \textbf{81.9} & 47.8          & 65.4          & \textbf{77.3} & \textbf{56.3} & 61.9         \\
\bottomrule
\end{tabular}
\caption{SU-TIS Main results on ToTTo and HiTab. We also report accuracy across different option generation strategies: EJ (Exam-Judge), MI (Modify-Input), MO (Modify-Output), HA (Human-Annotation). \label{tab:tis_tsu_all}}
\end{threeparttable}
\end{table*}

\renewcommand{\arraystretch}{1.4} 
\begin{table*}[]\centering
\fontsize{10}{10}\selectfont
\begin{threeparttable}
\begin{tabular}{@{}l|ccccc|ccccc@{}}
\toprule
\multicolumn{1}{l|}{\multirow{2}{*}{\textbf{Model}}} & \multicolumn{5}{c|}{\textbf{ToTTo}}   & \multicolumn{5}{c}{\textbf{HiTab}}                                            \\
\multicolumn{1}{l|}{}                                & \textbf{EJ}   & \textbf{MI}   & \textbf{MO}   & \textbf{HA}   & \textbf{Avg.} & \textbf{EJ}   & \textbf{MI}   & \textbf{MO}   & \textbf{HA}   & \textbf{Avg.} \\
\midrule
\textit{proprietary model}                          &               &               &               &               &               &               &               &               &               &               \\
gemini-pro                                          & 63.2          & 84.6          & 86.2          & 62.8          & 79.4          & 49.8          & 67.4          & 78.3          & 65.5          & 64.8          \\
GPT-3.5-turbo-instruct                              & 61.9          & 81.1          & 79.7          & 53.5          & 74.5          & 59.3          & 68.5          & 79.2          & 48.4          & 66.8          \\
GPT-3.5-turbo-1106                                  & 55.0          & 71.0          & 72.0          & 53.5          & 66.7          & 39.7          & 59.2          & 66.1          & 41.9          & 53.0          \\
GPT-3.5-turbo-16k                                   & 56.5          & 81.6          & 79.3          & 53.5          & 73.4          & 45.0          & 67.2          & 73.8          & 39.8          & 59.2          \\
GPT-4                                               & \textbf{74.2} & \textbf{93.1} & \textbf{96.2} & \textbf{86.1} & \textbf{89.7} & \textbf{71.0} & \textbf{83.1} & \textbf{91.4} & \textbf{72.0} & \textbf{80.4} \\
\midrule
\textit{open-source model}                                &               &               &               &               &               &               &               &               &               &               \\
Llama2-chat-7b                                      & 51.2          & 51.1          & 52.8          & 58.1          & 52.3          & 45.5          & 53.8          & 48.8          & 43.0          & 48.6          \\
TableLlama-7b                                       & 51.9          & 53.9          & 54.9          & 57.0          & 54.1          & 45.0          & 52.2          & 49.1          & 40.9          & 47.9          \\
Mistral-7b-instruct-v0.2                            & 46.9          & 74.2          & 76.7          & 66.3          & 68.8          & 41.9          & 63.7          & 71.4          & 47.3          & 57.1          \\
Llama2-chat-13b                                     & 52.3          & 63.5          & 63.0          & 57.0          & 60.5          & 44.2          & 53.2          & 67.0          & \textbf{55.9} & 54.4          \\
Mixtral-8*7b-instruct                               & 50.4          & 84.4          & 84.6          & 69.8          & 76.2          & 46.5          & 59.9          & 72.3          & 47.3          & 57.9          \\
Llama2-chat-70b                                     & 47.7          & 69.7          & 76.3          & 67.4          & 67.4          & 43.9          & 59.9          & 64.0          & 49.5          & 54.7          \\
Tulu2-70b-DPO      & \textbf{60.4} & \textbf{90.3} & \textbf{90.2} & \textbf{76.7} & \textbf{82.9} & \textbf{52.0} & \textbf{68.8} & \textbf{76.8} & 52.7          & \textbf{64.0} \\
\bottomrule
\end{tabular}
\caption{M-TIS Main results on ToTTo and HiTab. We also report accuracy across different option generation strategies: EJ (Exam-Judge), MI (Modify-Input), MO (Modify-Output), HA (Human-Annotation). \label{tab:tis_prt_all}}
\end{threeparttable}
\end{table*}

%% file: emnlp2023.bbl
\begin{thebibliography}{38}
\expandafter\ifx\csname natexlab\endcsname\relax\def\natexlab#1{#1}\fi

\bibitem[{Akhtar et~al.(2023)Akhtar, Shankarampeta, Gupta, Patil, Cocarascu, and Simperl}]{numerical_reasoning}
Mubashara Akhtar, Abhilash~Reddy Shankarampeta, Vivek Gupta, Arpit Patil, Oana Cocarascu, and Elena Simperl. 2023.
\newblock Exploring the numerical reasoning capabilities of language models: {A} comprehensive analysis on tabular data.
\newblock In \emph{Findings of the Association for Computational Linguistics: {EMNLP} 2023, Singapore, December 6-10, 2023}.

\bibitem[{Brown et~al.(2020)Brown, Mann, Ryder, Subbiah, Kaplan, Dhariwal, Neelakantan, Shyam, Sastry, Askell, Agarwal, Herbert{-}Voss, Krueger, Henighan, Child, Ramesh, Ziegler, Wu, Winter, Hesse, Chen, Sigler, Litwin, Gray, Chess, Clark, Berner, McCandlish, Radford, Sutskever, and Amodei}]{gpt3}
Tom~B. Brown, Benjamin Mann, Nick Ryder, Melanie Subbiah, Jared Kaplan, Prafulla Dhariwal, Arvind Neelakantan, Pranav Shyam, Girish Sastry, Amanda Askell, Sandhini Agarwal, Ariel Herbert{-}Voss, Gretchen Krueger, Tom Henighan, Rewon Child, Aditya Ramesh, Daniel~M. Ziegler, Jeffrey Wu, Clemens Winter, Christopher Hesse, Mark Chen, Eric Sigler, Mateusz Litwin, Scott Gray, Benjamin Chess, Jack Clark, Christopher Berner, Sam McCandlish, Alec Radford, Ilya Sutskever, and Dario Amodei. 2020.
\newblock \href {https://proceedings.neurips.cc/paper/2020/hash/1457c0d6bfcb4967418bfb8ac142f64a-Abstract.html} {Language models are few-shot learners}.
\newblock In \emph{Advances in Neural Information Processing Systems 33: Annual Conference on Neural Information Processing Systems 2020, NeurIPS 2020, December 6-12, 2020, virtual}.

\bibitem[{Chen et~al.(2020{\natexlab{a}})Chen, Chen, Su, Chen, and Wang}]{nli_acc}
Wenhu Chen, Jianshu Chen, Yunde Su, Zhiyu Chen, and William~Yang Wang. 2020{\natexlab{a}}.
\newblock Logical natural language generation from open-domain tables.
\newblock \emph{ArXiv}, abs/2004.10404.

\bibitem[{Chen et~al.(2020{\natexlab{b}})Chen, Zha, Chen, Xiong, Wang, and Wang}]{hybridqa}
Wenhu Chen, Hanwen Zha, Zhiyu Chen, Wenhan Xiong, Hong Wang, and William~Yang Wang. 2020{\natexlab{b}}.
\newblock Hybridqa: {A} dataset of multi-hop question answering over tabular and textual data.
\newblock In \emph{Findings of the Association for Computational Linguistics: {EMNLP} 2020, Online Event, 16-20 November 2020}.

\bibitem[{Cheng et~al.(2022)Cheng, Dong, Wang, Jia, Guo, Gao, Han, Lou, and Zhang}]{hitab}
Zhoujun Cheng, Haoyu Dong, Zhiruo Wang, Ran Jia, Jiaqi Guo, Yan Gao, Shi Han, Jian{-}Guang Lou, and Dongmei Zhang. 2022.
\newblock Hitab: {A} hierarchical table dataset for question answering and natural language generation.
\newblock In \emph{Proceedings of the 60th Annual Meeting of the Association for Computational Linguistics (Volume 1: Long Papers), {ACL} 2022, Dublin, Ireland, May 22-27, 2022}.

\bibitem[{Chowdhery et~al.(2022)Chowdhery, Narang, Devlin, Bosma, Mishra, Roberts, Barham, Chung, Sutton, Gehrmann, Schuh, Shi, Tsvyashchenko, Maynez, Rao, Barnes, Tay, Shazeer, Prabhakaran, Reif, Du, Hutchinson, Pope, Bradbury, Austin, Isard, Gur{-}Ari, Yin, Duke, Levskaya, Ghemawat, Dev, Michalewski, Garcia, Misra, Robinson, Fedus, Zhou, Ippolito, Luan, Lim, Zoph, Spiridonov, Sepassi, Dohan, Agrawal, Omernick, Dai, Pillai, Pellat, Lewkowycz, Moreira, Child, Polozov, Lee, Zhou, Wang, Saeta, Diaz, Firat, Catasta, Wei, Meier{-}Hellstern, Eck, Dean, Petrov, and Fiedel}]{palm}
Aakanksha Chowdhery, Sharan Narang, Jacob Devlin, Maarten Bosma, Gaurav Mishra, Adam Roberts, Paul Barham, Hyung~Won Chung, Charles Sutton, Sebastian Gehrmann, Parker Schuh, Kensen Shi, Sasha Tsvyashchenko, Joshua Maynez, Abhishek Rao, Parker Barnes, Yi~Tay, Noam Shazeer, Vinodkumar Prabhakaran, Emily Reif, Nan Du, Ben Hutchinson, Reiner Pope, James Bradbury, Jacob Austin, Michael Isard, Guy Gur{-}Ari, Pengcheng Yin, Toju Duke, Anselm Levskaya, Sanjay Ghemawat, Sunipa Dev, Henryk Michalewski, Xavier Garcia, Vedant Misra, Kevin Robinson, Liam Fedus, Denny Zhou, Daphne Ippolito, David Luan, Hyeontaek Lim, Barret Zoph, Alexander Spiridonov, Ryan Sepassi, David Dohan, Shivani Agrawal, Mark Omernick, Andrew~M. Dai, Thanumalayan~Sankaranarayana Pillai, Marie Pellat, Aitor Lewkowycz, Erica Moreira, Rewon Child, Oleksandr Polozov, Katherine Lee, Zongwei Zhou, Xuezhi Wang, Brennan Saeta, Mark Diaz, Orhan Firat, Michele Catasta, Jason Wei, Kathy Meier{-}Hellstern, Douglas Eck, Jeff Dean, Slav Petrov, and Noah Fiedel.
  2022.
\newblock \href {https://doi.org/10.48550/arXiv.2204.02311} {Palm: Scaling language modeling with pathways}.
\newblock \emph{CoRR}, abs/2204.02311.

\bibitem[{Dhingra et~al.(2019)Dhingra, Faruqui, Parikh, Chang, Das, and Cohen}]{parent}
Bhuwan Dhingra, Manaal Faruqui, Ankur~P. Parikh, Ming{-}Wei Chang, Dipanjan Das, and William~W. Cohen. 2019.
\newblock Handling divergent reference texts when evaluating table-to-text generation.
\newblock In \emph{Proceedings of the 57th Conference of the Association for Computational Linguistics, {ACL} 2019, Florence, Italy, July 28- August 2, 2019, Volume 1: Long Papers}.

\bibitem[{Google(2023)}]{gemini-pro}
Google. 2023.
\newblock \href {https://blog.google/technology/ai/google-gemini-ai} {Introducing gemini: our largest and most capable ai model}.

\bibitem[{Hendrycks et~al.(2020)Hendrycks, Burns, Basart, Zou, Mazeika, Song, and Steinhardt}]{mmlu}
Dan Hendrycks, Collin Burns, Steven Basart, Andy Zou, Mantas Mazeika, Dawn~Xiaodong Song, and Jacob Steinhardt. 2020.
\newblock \href {https://api.semanticscholar.org/CorpusID:221516475} {Measuring massive multitask language understanding}.
\newblock \emph{ArXiv}, abs/2009.03300.

\bibitem[{hiyouga(2023)}]{llama-factory}
hiyouga. 2023.
\newblock Llama factory.
\newblock \url{https://github.com/hiyouga/LLaMA-Factory}.

\bibitem[{Ivison et~al.(2023)Ivison, Wang, Pyatkin, Lambert, Peters, Dasigi, Jang, Wadden, Smith, Beltagy, and Hajishirzi}]{tulu2}
Hamish Ivison, Yizhong Wang, Valentina Pyatkin, Nathan Lambert, Matthew Peters, Pradeep Dasigi, Joel Jang, David Wadden, Noah~A. Smith, Iz~Beltagy, and Hannaneh Hajishirzi. 2023.
\newblock Camels in a changing climate: Enhancing {LM} adaptation with tulu 2.
\newblock \emph{CoRR}, abs/2311.10702.

\bibitem[{Jiang et~al.(2023)Jiang, Sablayrolles, Mensch, Bamford, Chaplot, de~las Casas, Bressand, Lengyel, Lample, Saulnier, Lavaud, Lachaux, Stock, Scao, Lavril, Wang, Lacroix, and Sayed}]{mistral-7b}
Albert~Q. Jiang, Alexandre Sablayrolles, Arthur Mensch, Chris Bamford, Devendra~Singh Chaplot, Diego de~las Casas, Florian Bressand, Gianna Lengyel, Guillaume Lample, Lucile Saulnier, Lélio~Renard Lavaud, Marie-Anne Lachaux, Pierre Stock, Teven~Le Scao, Thibaut Lavril, Thomas Wang, Timothée Lacroix, and William~El Sayed. 2023.
\newblock \href {http://arxiv.org/abs/2310.06825} {Mistral 7b}.

\bibitem[{Jiang et~al.(2024)Jiang, Sablayrolles, Roux, Mensch, Savary, Bamford, Chaplot, de~Las~Casas, Hanna, Bressand, Lengyel, Bour, Lample, Lavaud, Saulnier, Lachaux, Stock, Subramanian, Yang, Antoniak, Scao, Gervet, Lavril, Wang, Lacroix, and Sayed}]{mixtral-8*7b}
Albert~Q. Jiang, Alexandre Sablayrolles, Antoine Roux, Arthur Mensch, Blanche Savary, Chris Bamford, Devendra~Singh Chaplot, Diego de~Las~Casas, Emma~Bou Hanna, Florian Bressand, Gianna Lengyel, Guillaume Bour, Guillaume Lample, L'elio~Renard Lavaud, Lucile Saulnier, Marie-Anne Lachaux, Pierre Stock, Sandeep Subramanian, Sophia Yang, Szymon Antoniak, Teven~Le Scao, Th{\'e}ophile Gervet, Thibaut Lavril, Thomas Wang, Timoth{\'e}e Lacroix, and William~El Sayed. 2024.
\newblock \href {https://api.semanticscholar.org/CorpusID:266844877} {Mixtral of experts}.

\bibitem[{Lebret et~al.(2016)Lebret, Grangier, and Auli}]{app_biograph}
R{\'e}mi Lebret, David Grangier, and Michael Auli. 2016.
\newblock Neural text generation from structured data with application to the biography domain.
\newblock In \emph{Conference on Empirical Methods in Natural Language Processing}.

\bibitem[{Lehmberg et~al.(2016)Lehmberg, Ritze, Meusel, and Bizer}]{lehmberg2016large}
Oliver Lehmberg, Dominique Ritze, Robert Meusel, and Christian Bizer. 2016.
\newblock A large public corpus of web tables containing time and context metadata.
\newblock In \emph{Proceedings of the 25th international conference companion on world wide web}, pages 75--76.

\bibitem[{Lin(2004)}]{rouge}
Chin-Yew Lin. 2004.
\newblock Rouge: A package for automatic evaluation of summaries.
\newblock In \emph{Annual Meeting of the Association for Computational Linguistics}.

\bibitem[{Liu et~al.(2023)Liu, Lin, Hewitt, Paranjape, Bevilacqua, Petroni, and Liang}]{lost_in_the_middle}
Nelson~F. Liu, Kevin Lin, John Hewitt, Ashwin Paranjape, Michele Bevilacqua, Fabio Petroni, and Percy Liang. 2023.
\newblock Lost in the middle: How language models use long contexts.
\newblock \emph{CoRR}, abs/2307.03172.

\bibitem[{Nan et~al.(2021)Nan, Hsieh, Mao, Lin, Verma, Zhang, Kryscinski, Schoelkopf, Kong, Tang, Mutuma, Rosand, Trindade, Bandaru, Cunningham, Xiong, and Radev}]{fetaqa}
Linyong Nan, Chiachun Hsieh, Ziming Mao, Xi~Victoria Lin, Neha Verma, Rui Zhang, Wojciech Kryscinski, Hailey Schoelkopf, Riley Kong, Xiangru Tang, Mutethia Mutuma, Ben Rosand, Isabel Trindade, Renusree Bandaru, Jacob Cunningham, Caiming Xiong, and Dragomir~R. Radev. 2021.
\newblock Fetaqa: Free-form table question answering.
\newblock \emph{Trans. Assoc. Comput. Linguistics}, 10:35--49.

\bibitem[{OpenAI(2023{\natexlab{a}})}]{gpt4}
OpenAI. 2023{\natexlab{a}}.
\newblock \href {https://api.semanticscholar.org/CorpusID:257532815} {Gpt-4 technical report}.

\bibitem[{OpenAI(2023{\natexlab{b}})}]{chatgpt}
OpenAI. 2023{\natexlab{b}}.
\newblock \href {https://openai.com/blog/chatgpt} {Introducing chatgpt.}

\bibitem[{Papineni et~al.(2002)Papineni, Roukos, Ward, and Zhu}]{bleu}
Kishore Papineni, Salim Roukos, Todd Ward, and Wei{-}Jing Zhu. 2002.
\newblock Bleu: a method for automatic evaluation of machine translation.
\newblock In \emph{Proceedings of the 40th Annual Meeting of the Association for Computational Linguistics, July 6-12, 2002, Philadelphia, PA, {USA}}.

\bibitem[{Parikh et~al.(2020)Parikh, Wang, Gehrmann, Faruqui, Dhingra, Yang, and Das}]{totto}
Ankur~P. Parikh, Xuezhi Wang, Sebastian Gehrmann, Manaal Faruqui, Bhuwan Dhingra, Diyi Yang, and Dipanjan Das. 2020.
\newblock Totto: {A} controlled table-to-text generation dataset.
\newblock In \emph{Proceedings of the 2020 Conference on Empirical Methods in Natural Language Processing, {EMNLP} 2020, Online, November 16-20, 2020}.

\bibitem[{Pasupat and Liang(2015)}]{wikitablequestions}
Panupong Pasupat and Percy Liang. 2015.
\newblock Compositional semantic parsing on semi-structured tables.
\newblock In \emph{Proceedings of the 53rd Annual Meeting of the Association for Computational Linguistics and the 7th International Joint Conference on Natural Language Processing of the Asian Federation of Natural Language Processing, {ACL} 2015, July 26-31, 2015, Beijing, China, Volume 1: Long Papers}.

\bibitem[{Rafailov et~al.(2023)Rafailov, Sharma, Mitchell, Ermon, Manning, and Finn}]{dpo}
Rafael Rafailov, Archit Sharma, Eric Mitchell, Stefano Ermon, Christopher~D. Manning, and Chelsea Finn. 2023.
\newblock Direct preference optimization: Your language model is secretly a reward model.
\newblock \emph{ArXiv}.

\bibitem[{Rajbhandari et~al.(2020)Rajbhandari, Rasley, Ruwase, and He}]{zero}
Samyam Rajbhandari, Jeff Rasley, Olatunji Ruwase, and Yuxiong He. 2020.
\newblock Zero: memory optimizations toward training trillion parameter models.
\newblock In \emph{Proceedings of the International Conference for High Performance Computing, Networking, Storage and Analysis, {SC} 2020, Virtual Event / Atlanta, Georgia, USA, November 9-19, 2020}, page~20. {IEEE/ACM}.

\bibitem[{Sui et~al.(2024)Sui, Zhou, Zhou, Han, and Zhang}]{gpt4table}
Yuan Sui, Mengyu Zhou, Mingjie Zhou, Shi Han, and Dongmei Zhang. 2024.
\newblock Gpt4table: Can large language models understand structured table data? a benchmark and empirical study.
\newblock In \emph{WSDM 2024}.

\bibitem[{Suzgun et~al.(2022)Suzgun, Scales, Scharli, Gehrmann, Tay, Chung, Chowdhery, Le, hsin Chi, Zhou, and Wei}]{bbh}
Mirac Suzgun, Nathan Scales, Nathanael Scharli, Sebastian Gehrmann, Yi~Tay, Hyung~Won Chung, Aakanksha Chowdhery, Quoc~V. Le, Ed~Huai hsin Chi, Denny Zhou, and Jason Wei. 2022.
\newblock \href {https://api.semanticscholar.org/CorpusID:252917648} {Challenging big-bench tasks and whether chain-of-thought can solve them}.
\newblock In \emph{Annual Meeting of the Association for Computational Linguistics}.

\bibitem[{Touvron et~al.(2023)Touvron, Martin, Stone, Albert, Almahairi, Babaei, Bashlykov, Batra, Bhargava, Bhosale, Bikel, Blecher, Ferrer, Chen, Cucurull, Esiobu, Fernandes, Fu, Fu, Fuller, Gao, Goswami, Goyal, Hartshorn, Hosseini, Hou, Inan, Kardas, Kerkez, Khabsa, Kloumann, Korenev, Koura, Lachaux, Lavril, Lee, Liskovich, Lu, Mao, Martinet, Mihaylov, Mishra, Molybog, Nie, Poulton, Reizenstein, Rungta, Saladi, Schelten, Silva, Smith, Subramanian, Tan, Tang, Taylor, Williams, Kuan, Xu, Yan, Zarov, Zhang, Fan, Kambadur, Narang, Rodriguez, Stojnic, Edunov, and Scialom}]{llama2}
Hugo Touvron, Louis Martin, Kevin~R. Stone, Peter Albert, Amjad Almahairi, Yasmine Babaei, Nikolay Bashlykov, Soumya Batra, Prajjwal Bhargava, Shruti Bhosale, Daniel~M. Bikel, Lukas Blecher, Cristian~Cant{\'o}n Ferrer, Moya Chen, Guillem Cucurull, David Esiobu, Jude Fernandes, Jeremy Fu, Wenyin Fu, Brian Fuller, Cynthia Gao, Vedanuj Goswami, Naman Goyal, Anthony~S. Hartshorn, Saghar Hosseini, Rui Hou, Hakan Inan, Marcin Kardas, Viktor Kerkez, Madian Khabsa, Isabel~M. Kloumann, A.~V. Korenev, Punit~Singh Koura, Marie-Anne Lachaux, Thibaut Lavril, Jenya Lee, Diana Liskovich, Yinghai Lu, Yuning Mao, Xavier Martinet, Todor Mihaylov, Pushkar Mishra, Igor Molybog, Yixin Nie, Andrew Poulton, Jeremy Reizenstein, Rashi Rungta, Kalyan Saladi, Alan Schelten, Ruan Silva, Eric~Michael Smith, R.~Subramanian, Xia Tan, Binh Tang, Ross Taylor, Adina Williams, Jian~Xiang Kuan, Puxin Xu, Zhengxu Yan, Iliyan Zarov, Yuchen Zhang, Angela Fan, Melanie Kambadur, Sharan Narang, Aurelien Rodriguez, Robert Stojnic, Sergey Edunov, and
  Thomas Scialom. 2023.
\newblock Llama 2: Open foundation and fine-tuned chat models.
\newblock \emph{ArXiv}, abs/2307.09288.

\bibitem[{Wang et~al.(2022)Wang, Xu, Szekely, and Chen}]{lattice}
Fei Wang, Zhewei Xu, Pedro~A. Szekely, and Muhao Chen. 2022.
\newblock Robust (controlled) table-to-text generation with structure-aware equivariance learning.
\newblock \emph{ArXiv}.

\bibitem[{Wang et~al.(2023)Wang, Ivison, Dasigi, Hessel, Khot, Chandu, Wadden, MacMillan, Smith, Beltagy, and Hajishirzi}]{tulu}
Yizhong Wang, Hamish Ivison, Pradeep Dasigi, Jack Hessel, Tushar Khot, Khyathi~Raghavi Chandu, David Wadden, Kelsey MacMillan, Noah~A. Smith, Iz~Beltagy, and Hannaneh Hajishirzi. 2023.
\newblock \href {https://doi.org/10.48550/ARXIV.2306.04751} {How far can camels go? exploring the state of instruction tuning on open resources}.
\newblock \emph{CoRR}, abs/2306.04751.

\bibitem[{Wang et~al.(2020)Wang, Dong, Jia, Li, Fu, Han, and Zhang}]{wang2020structure}
Zhiruo Wang, Haoyu Dong, Ran Jia, Jia Li, Zhiyi Fu, Shi Han, and Dongmei Zhang. 2020.
\newblock Structure-aware pre-training for table understanding with tree-based transformers.
\newblock \emph{arXiv preprint arXiv:2010.12537}.

\bibitem[{Wiseman et~al.(2017)Wiseman, Shieber, and Rush}]{app_sport}
Sam Wiseman, Stuart~M. Shieber, and Alexander~M. Rush. 2017.
\newblock Challenges in data-to-document generation.
\newblock \emph{ArXiv}.

\bibitem[{Yang et~al.(2023)Yang, Tang, Zhao, Xiao, and Lin}]{distill_table_reason}
Bohao Yang, Chen Tang, Kun Zhao, Chenghao Xiao, and Chenghua Lin. 2023.
\newblock Effective distillation of table-based reasoning ability from llms.
\newblock \emph{CoRR}, abs/2309.13182.

\bibitem[{Zeng et~al.(2023)Zeng, Yu, Gao, Meng, Goyal, and Chen}]{evaleval}
Zhiyuan Zeng, Jiatong Yu, Tianyu Gao, Yu~Meng, Tanya Goyal, and Danqi Chen. 2023.
\newblock Evaluating large language models at evaluating instruction following.
\newblock \emph{CoRR}, abs/2310.07641.

\bibitem[{Zhang et~al.(2023)Zhang, Yue, Li, and Sun}]{tablellama}
Tianshu Zhang, Xiang Yue, Yifei Li, and Huan Sun. 2023.
\newblock Tablellama: Towards open large generalist models for tables.
\newblock \emph{CoRR}, abs/2311.09206.

\bibitem[{Zhao et~al.(2023{\natexlab{a}})Zhao, Mi, Qi, Nan, Guo, Cohan, and Radev}]{openrt}
Yilun Zhao, Boyu Mi, Zhenting Qi, Linyong Nan, Minghao Guo, Arman Cohan, and Dragomir~R. Radev. 2023{\natexlab{a}}.
\newblock Openrt: An open-source framework for reasoning over tabular data.
\newblock In \emph{Annual Meeting of the Association for Computational Linguistics}.

\bibitem[{Zhao et~al.(2023{\natexlab{b}})Zhao, Zhang, Si, Nan, Tang, and Cohan}]{invest_ttg}
Yilun Zhao, Haowei Zhang, Shengyun Si, Linyong Nan, Xiangru Tang, and Arman Cohan. 2023{\natexlab{b}}.
\newblock Investigating table-to-text generation capabilities of large language models in real-world information seeking scenarios.
\newblock In \emph{Conference on Empirical Methods in Natural Language Processing}.

\bibitem[{Zhong et~al.(2017)Zhong, Xiong, and Socher}]{wikisql}
Victor Zhong, Caiming Xiong, and Richard Socher. 2017.
\newblock Seq2sql: Generating structured queries from natural language using reinforcement learning.
\newblock \emph{CoRR}, abs/1709.00103.

\end{thebibliography}
